\documentclass[runningheads]{llncs}

\usepackage{eccv}

\usepackage{eccvabbrv}

\usepackage{graphicx}
\usepackage{booktabs}

\usepackage[accsupp]{axessibility}  

\usepackage[pagebackref,breaklinks,colorlinks,citecolor=eccvblue]{hyperref}

\usepackage{orcidlink}
\usepackage{caption}
\usepackage{colortbl}
\usepackage{tabularx}
\usepackage{makecell}
\usepackage{multirow}
\usepackage{tikz}
\usepackage{times}
\usepackage{epsfig}
\usepackage{amsmath}
\usepackage{amssymb}
\usepackage{amssymb}
\usepackage{pifont}
\newcommand{\cmark}{\ding{51}}%
\newcommand{\xmark}{\ding{55}}%
\usepackage{algorithm}
\usepackage{algpseudocode}
\usepackage{booktabs}
\usepackage{xcolor}
\usepackage{multicol}
\usepackage{diagbox}
\begin{document}

\title{Unified Batch Normalization: Identifying and Alleviating the Feature Condensation in Batch Normalization and a Unified Framework}
\titlerunning{Unified Batch Normalization}

\author{Shaobo Wang \and Xiangdong Zhang \and Dongrui Liu \and Junchi Yan\thanks{Correspondence to: Junchi Yan (yanjunchi@sjtu.edu.cn)}}

\authorrunning{Shaobo Wang et al.}

\institute{Shanghai Jiao Tong University, Shanghai, China
\email{\{shaobowang1009,zhangxiangdong,drliu96,yanjunchi\}@sjtu.edu.cn}}

\maketitle

\begin{abstract}
Batch Normalization (BN) has become an essential technique in contemporary neural network design, enhancing training stability. Specifically, BN employs centering and scaling operations to standardize features along the batch dimension and uses an affine transformation to recover features. Although standard BN has shown its capability to improve deep neural network training and convergence, it still exhibits inherent limitations in certain cases. Current enhancements to BN typically address only isolated aspects of its mechanism. In this work, we critically examine BN from a feature perspective, identifying feature condensation during BN as a detrimental factor to test performance. To tackle this problem, we propose a two-stage unified framework called Unified Batch Normalization (UBN). In the first stage, we employ a straightforward feature condensation threshold to mitigate condensation effects, thereby preventing improper updates of statistical norms. In the second stage, we unify various normalization variants to boost each component of BN. Our experimental results reveal that UBN significantly enhances performance across different visual backbones and different vision tasks, and notably expedites network training convergence, particularly in early training stages. Notably, our method improved about 3\% in accuracy on ImageNet classification and 4\% in mean average precision on both Object Detection and Instance Segmentation on COCO dataset, showing the effectiveness of our approach in real-world scenarios. 
\end{abstract}

\section{Introduction}
\label{sec:intro}
Deep neural networks (DNNs) have made remarkable strides in enhancing computer vision performance, primarily attributed to their extraordinary representational capabilities. Among various DNNs, Convolutional Neural Networks (CNNs) serve as backbones for feature extraction in downstream tasks. Batch Normalization (BN) \cite{ioffe2015batch} is a widely adopted technique in the training of CNNs, which enhances the stability of optimization through a combination of feature normalization and affine transformation.

\begin{table}[tb!]
\caption{Study on the rectifications of the three components widely used in BN. Classical normalization variants only address a single component of BN. In comparison, our proposed UBN unifies diverse normalization variants and alleviates feature condensation to perform rectifications in all aspects of BN, thereby enhancing the testing performance and training efficiency.}
\vspace{-10pt}
\label{table:comparison}
\centering
\begin{tabular}{@{}ccccccc@{}}
\toprule
Component & BN \cite{ioffe2015batch} & BNET \cite{10012548} & IEBN \cite{liang2019instance} & RBN \cite{gao2021representative}& UBN (ours) \\ 
\midrule
Normalization & \xmark & \xmark & \cmark & \cmark & \cmark \\
Affine & \xmark & \cmark & \xmark & \xmark & \cmark \\
Running statistics & \xmark & \xmark & \xmark & \xmark & \cmark \\
\bottomrule
\end{tabular}
\vspace{-15pt}
\end{table}

However, in recent years, shortcomings of the BN have been witnessed in some specific applications \cite{wang2022understanding, bjorck2018understanding, thakkar2018batch, peng2018megdet, faster2015towards}. Table~\ref{table:comparison} shows some representative normalization works. Some variants of BN  focus on normalization rectification, which involves centering or scaling rectification. For centering rectification, most papers focus on rectifying the input features, \emph{i.e.}, partitioning input features into different groups or different dimensions \cite{ulyanov2017instance, pan2018two, wu2018group, li2019positional}, which may not fully utilize batch statistics. For scaling rectification, \cite{gao2021representative} tackled the intensity of scaling with additional modification of features, \emph{e.g.}, by leveraging multi-step statistics, or multi-batch statistics \cite{guo2020double,zhu2020CBN}. Besides, \cite{10012548} explores how to improve affine transformation with local channel statistics. In fact, these variants predominantly concentrate on individual or a few aspects within BN without offering a comprehensive solution to rectify the inconsistency in statistics across all components. This limitation has resulted in restricted practical applicability.


To overcome these challenges, we propose a comprehensive, two-stage approach named Unified Batch Normalization (UBN). Our initial analysis of BN from a feature learning perspective uncovers a \textit{feature condensation phenomenon} within DNNs utilizing BN, which poses a barrier to effective learning. We ascribe BN's limitations to its ineffectiveness in counteracting feature condensation. In response, our first stage introduces a simple yet impactful ``Feature Condensation Threshold,'' guiding the selection of statistics during training. The second stage presents a unified framework to enhance BN, incorporating rectifications across its components to improve overall performance. Our contributions are as follows:




\begin{enumerate}
    \item To our best knowledge, we are the first to identify the \textit{feature condensation phenomenon} in BN, a critical insight that sheds light on previously unaddressed impediments to the learning of normalization methods.
    \item We introduce an innovative conditional update strategy with the introduction of \textit{feature condensation threshold.} This approach selectively navigates the normalization statistics during training, effectively mitigating the feature condensation issue and promoting more dynamic and adaptable learning pathways.
    \item We propose a unified framework that unifies typical rectifications of components in BN \cite{gao2021representative,10012548,liang2019instance}, which can be incorporated into different training scenarios.
    \item We evaluate UBN's effectiveness across different image datasets and vision tasks, showcasing notable testing accuracy improvements (\emph{e.g.}, exceeding 3\% accuracy gain on ImageNet)  and early-stage training efficiency boosts. UBN also achieves significant average precision improvements (\emph{e.g.}, exceeding 4\% mean average precision gains on COCO) in both instance segmentation and object detection tasks.
\end{enumerate}

\begin{figure}[tb!]
    \centering
    \includegraphics[width=0.99\linewidth]{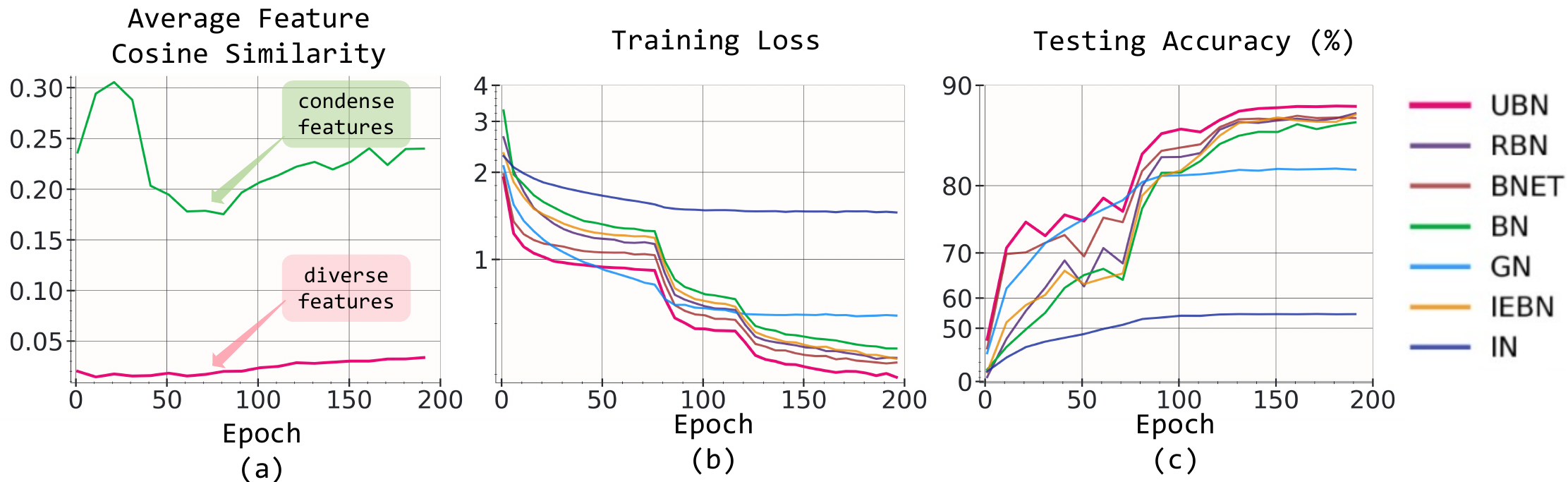}
    \captionsetup{skip=3pt}
    \caption{The feature condensation phenomenon and learning dynamics when performing normalization methods on the CIFAR-10 dataset (a) The average feature cosine similarity between input features of the first normalization layers in ResNet-34 \cite{he2015deep}. UBN reduces the feature condensation phenomenon by properly leveraging the running statistics, improving normalization performance. (b) The training loss of ResNet-34 of different representative normalization methods. (c) The testing accuracy of ResNet-34 of different representative normalization methods. UBN shows faster convergence and better performance than BN and other state-of-the-art (SOTA) normalization methods. We used non-uniform axis scale to visualize the differences between UBN and other SOTA normalization methods in terms of loss and accuracy.
    }
    \label{fig:ubn}
\vspace{-10pt}
\end{figure}


\section{Related Work}
\label{sec:related-work}
\subsection{Normalization Methods}     

Normalization techniques have played a crucial role in shaping the trajectory of neural network models \cite{chiley2019online,huang2023normalization,bulo2018place,de2017modulating,yuan2019generalized,cooijmans2016recurrent}. Among these variants, Batch Normalization (BN) \cite{ioffe2015batch} has markedly expedited training by global computing mean and variance across the batch dimension. The corpus of related work on BN predominantly bifurcates into two distinct streams. The first examines methods to refine BN's efficacy through modifications of its components. The second stream investigates strategies to derive more robust statistics for BN during training.
\vspace{-10pt}
\subsubsection{Component-Specific Approaches to Boost BN} Numerous studies have delved into specific components of BN, aiming to enhance or rectify individual aspects such as centering, scaling, and affine operations. Layer Normalization (LN) \cite{ba2016layer}, Group Normalization (GN) \cite{wu2018group}, Instance Normalization (IN) \cite{ulyanov2017instance}, and Positional Normalization \cite{li2019positional} perform different ways of partitioning the input features before centering to leverage statistics from other dimension/position/groups. Representative Batch Normalization (RBN) \cite{gao2021representative} focuses on centering and scaling adjustments to capture the representative features. BNET \cite{10012548} rectified the affine operation into a convolutional operation in CNNs to combine local channel information. Recent endeavors have also attempted to amalgamate multiple normalization approaches \cite{SwitchableNorm, shao2019ssn}. Nonetheless, these methodologies predominantly focus on singular or limited aspects of BN, not holistically enhancing its overall performance. Contrarily, our approach presents an integrated framework that amalgamates the best practices in normalization, offering a more comprehensive enhancement to BN's efficacy.

\vspace{-10pt}
\subsubsection{Running Statistics in BN across iterations}
Another focal point in BN research is optimizing running statistics over multiple batches and iterations. Considering the difference between training and inference, several studies have scrutinized the challenges associated with accumulated running statistics. Batch Re-Normalization \cite{ioffe2017batch} endeavored to enable the functionality of BN during training with small mini-batches. Memorized Batch Normalization \cite{guo2020double} took into account information from multiple recent batches to derive more precise and resilient statistics. AdaBN \cite{li2016revisiting} dynamically adjusted the statistics across all Batch Normalization layers, aiming to achieve profound adaptation effects in the context of domain adaptation tasks. Cross-Iteration Batch Normalization \cite{zhu2020CBN}  aggregates statistics across multiple training iterations to alleviate the mini-batch dependency. However, these approaches primarily consider neighboring statistics (such as neighboring batches or iterations), potentially neglecting a wider perspective of global running statistics. In contrast, our methodology proposes a novel approach to manage BN's global statistics by modulating feature condensation, thus offering a more versatile application across various normalization variants.

\subsection{Feature Learning}

In the context of feature learning, the examination of feature cosine similarity has played a pivotal role in understanding the feature condensation phenomenon, where features exhibit high cosine similarity. This research area has motivated our work, which centers on determining the running statistics in learning by leveraging feature condensation thresholds computed through feature cosine similarity.

Prior research in this domain has yielded valuable insights and techniques. \cite{siddharth2017learning} introduced disentangled feature learning techniques that can address the challenge of feature condensation. On the performance front, Cosine Normalization \cite{luo2018cosine} and works on Contrastive Learning \cite{chen2020simple,caron2021unsupervised} have harnessed feature cosine similarity to reduce redundancy and enhance feature representations. \cite{Zhao_2023_WACV} proposed dataset condensation to replace the original large training set with a significantly smaller learned set without neglecting importance statistics. \cite{liu2022trap} discussed a possible reason for the decrease of the feature diversity in the early learning of MLPs without batch normalization. \cite{zhou2023understanding} tried to explain the condensation of weights in CNNs in the initial learning phase. \cite{zhou2022batch} claimed that BN blocks the first and second derivatives \emph{w.r.t} loss functions and affects the feature-learning process in a negative way. Nevertheless, it is worth noting that while these methods have made strides in analyzing dynamics from a condensation perspective, they may need to explicitly tackle the challenge of feature condensation within BN. This limitation motivates our work to explore novel strategies for incorporating feature cosine similarity into BN to mitigate the effects of feature condensation, potentially improving convergence and overall model performance.

\begin{figure}[tb!]
    \centering
    \includegraphics[width=0.7\linewidth]{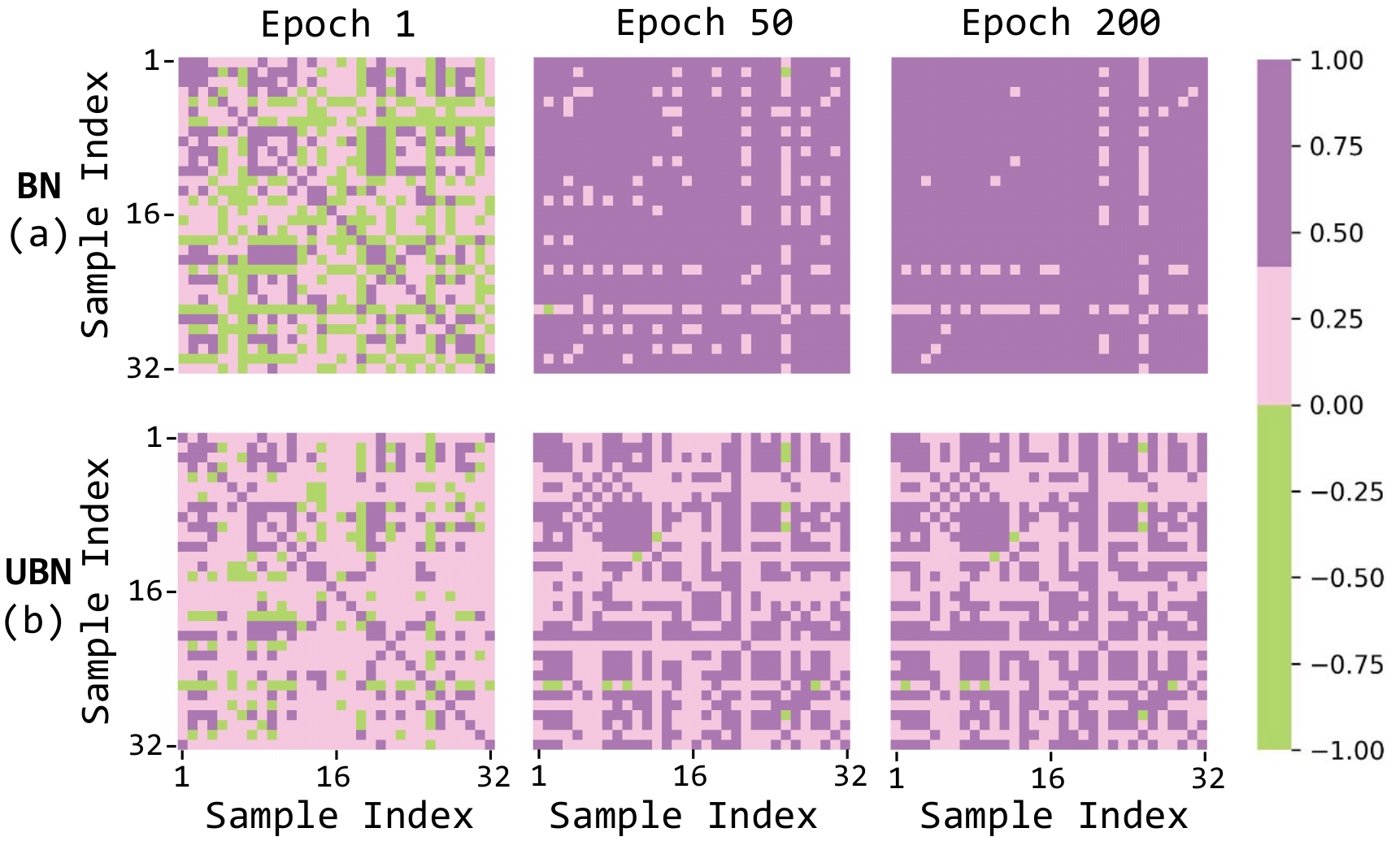}
    \caption{The feature condensation heatmap of ResNet-50 on the CIFAR-100 dataset. We randomly sampled 32 images as input and calculated the average cosine similarity of the input features of the first normalization layer of ResNet-50 before, during, and at the end of training. (a) When leveraging BN, the feature condensation became significant during training. (b) When leveraging UBN, the feature condensation phenomenon could be dramatically alleviated compared to BN. Note that the weight initialization of each model randomly decides the feature in the first epoch. 
    }
    \label{fig:sim}
\end{figure}

\section{Method}
\label{sec:method}
\subsection{Revisiting Batch Normalization}
We first revisit the formulation of Batch Normalization (BN). The standard BN comprises three components, \emph{i.e.}, centering, scaling, and affine transformation. In the computer vision task,  we denote a batch of input features as $\mathbf{X} \in \mathbb{R}^{B\times C\times H\times W}$, where $B, C, H,$ and $W$ are batch size, the number of channels, height, width of the input, respectively. BN normalizes each channel of input features in the given mini-batch into zero mean and unit variance. The formulation of BN is shown in Eq.~(\ref{eq:BN}).
\begin{equation}
    \label{eq:BN}
    \begin{aligned}
        & \text{Centering:} \ \mathbf{\bar{X}} = \mathbf{X} - \mu, \\
        & \text{Scaling:} \ \mathbf{\hat{X}} = \frac{\mathbf{\bar{X}}}{\sqrt{\sigma^2 + \varepsilon}}, \\
        & \text{Affine:} \ \mathbf{Y} = \mathbf{\mathbf{\hat{X}}} \gamma + \beta,
    \end{aligned}
\end{equation}
where $\mu, \sigma^2$ denote the mean and variance used for normalization, $\varepsilon$ is a small positive constant, and $\gamma, \beta$ denote the learned weight and bias used for affine transformation. During training, BN leverages the batch statistics for centering and scaling; during inference, the running mean and variance are used for normalization. The statistics used for normalization are shown in Eq.~(\ref{eq:bn_norm}).
\begin{equation}
    \label{eq:bn_norm}
    \begin{aligned}
            & \mu \gets \left\{
    \begin{aligned}
     m\mathbb{E}[\mathbf{X}] & +(1-m)\mu_{\text{old}}  & \ \text{(in training)}  \\
       & \mu_{\text{old}} & \ \text{(in inference)}  \\
    \end{aligned}
    \right. \\
    & \sigma^2 \gets \left\{
    \begin{aligned}
    m\text{Var}[\mathbf{X}]& +(1-m)\sigma^2_{\text{old}}   & \ \text{(in training)}  \\
    & \sigma^2_{\text{old}} & \ \text{(in inference)}  \\
    \end{aligned}
    \right.
    \end{aligned}
\end{equation}
While the mini-batch and running statistics are challenging to align strictly, testing instances may not consistently conform to the running distribution accumulated during training. This inconsistency between the training and inference processes weakens the effectiveness of BN \cite{yang2019mean}.

\subsection{Identifying the Feature Condensation Phenomenon in Training with BN}
Previous work mainly focused on improving a single or few components of BN. In contrast, we focus on a more fundamental problem in BN: \textit{how to determine the statistics in centering and scaling operations?} We aim to analyze it from a feature perspective. We assume that simply determining the normalization statistics according to the training or inference phase is inappropriate in BN because it ignores the difference between samples in each batch. Therefore, we aim to investigate the relationship between the input feature of BN and its relationship with performance. We formally define the feature condensation phenomenon in BN as follows:
\begin{definition}[Feature Condensation]
\label{def:fc}
In high-dimensional feature spaces, feature condensation is identified when a set of features \(X\) demonstrates an average cosine similarity $\mathbb{E}_{x,x'\in X,x\neq x'}\left[\cos(x,x')\right]$ exceeding a threshold \(\tau\).
\end{definition}

We first explore the learning dynamics with BN, and observe a notable pattern as depicted in Figure~\ref{fig:sim}. Specifically, feature similarity remains significantly elevated under BN. This phenomenon, termed feature condensation (Definition \ref{def:fc}), plays a pivotal role in the learning process. We observe that when the feature condensation remains to be high, the learning gets slow and eventually causes  a lower testing performance. 

The underlying cause of BN-induced feature condensation appears to be its normalization strategy, which centralizes means and minimizes variance among feature vectors {\small $x_i$} and {\small $x_j$}. This approach tends to elevate their dot product in relation to their norms, a process stabilized by BN. Hence, this standardized operation increases the average cosine similarity among feature vectors post-BN, often surpassing the threshold {\small $\tau$} and leading to feature condensation as defined. To tackle this problem, we adopt a simple yet effective threshold to set the statistics for normalization, as detailed in Section \ref{FCT}.

\subsection{Alleviating Feature Condensation with Rectifications}
Motivated by the limitations of Batch Normalization in scenarios characterized by feature condensation, our goal is to identify the most effective statistics for normalization. We aim to enhance the precision of the statistics used for centering and scaling, improve the  affine transformation recovery process, and retain the core advantages of BN. Our proposed solution, Unified Batch Normalization (UBN), specifically improves the batch statistics for targeted features, leading to gains in performance and increased reliability during the training of DNNs.


\subsubsection{Feature Condensation Rectification}
\label{FCT}
Previously, we attributed one main drawback of BN to its failure to effectively alleviate the feature condensation problem. We propose a simple yet effective threshold called Feature Condensation Threshold (FCT) to alleviate the feature condensation phenomenon better. According to Eq.~(\ref{eq:bn_norm}), BN uses different mean and variance during training and inference. However, different batches of features may share different distributions. In some cases, the feature similarity in a batch would be high; therefore, using centering and scaling operations would benefit the learning by reducing the similarity between instances; by contrast, when the feature similarity is already low, using the accumulated running mean and variance would stabilize training.

\begin{figure}[tb!]
    \centering
    \includegraphics[width=0.99\linewidth]{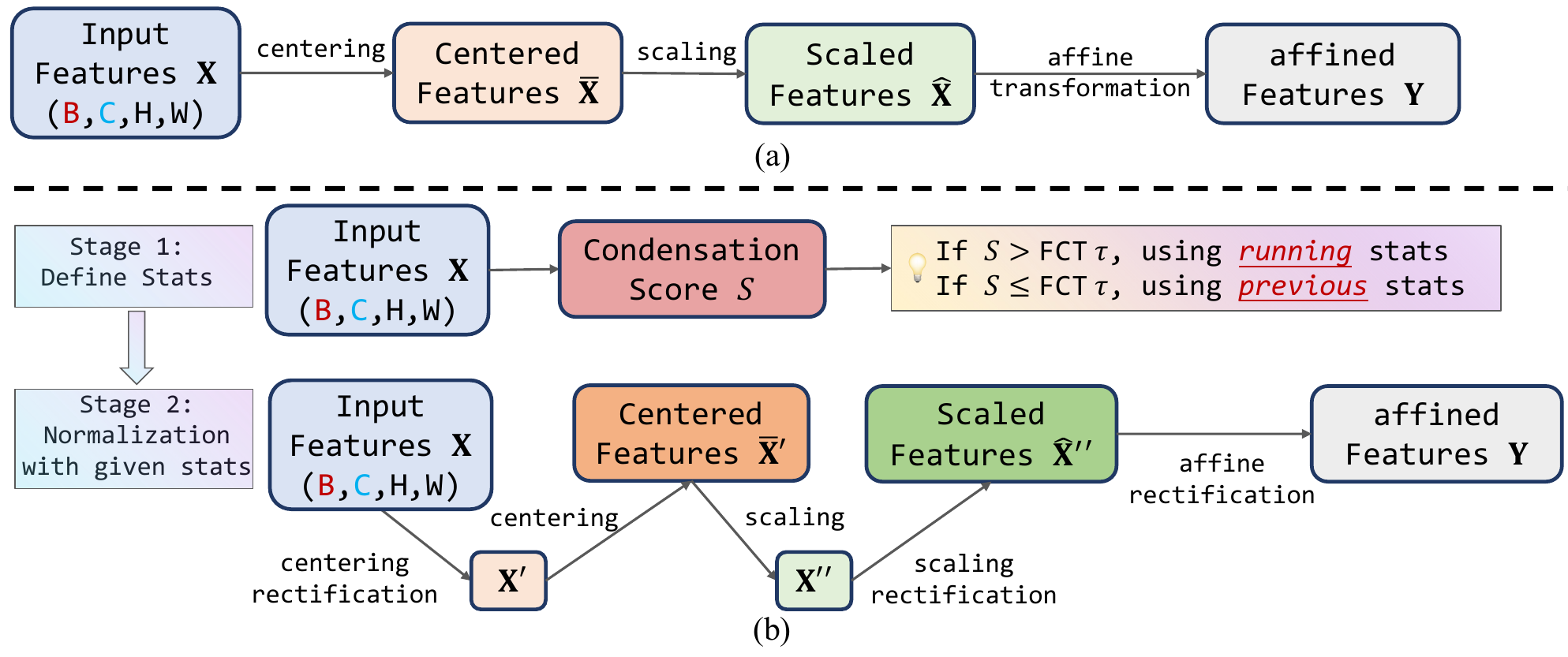}
    \captionsetup{skip=3pt}
    \caption{Sketch of the standard BN and our UBN. (a) BN consists of three components, \emph{i.e.}, centering, scaling, and affine transformation. (b) UBN improves BN with a two-stage method. In Stage 1, UBN defines the statistics for computing normalization by comparing the condensation score $S$ of the input features $\mathbf{X}$ with a given feature condensation threshold $\tau$. In Stage 2, UBN performs rectifications on each component of BN with the given statistics determined in Stage 1.
    }
    \label{fig:ubn}
\vspace{-10pt}
\end{figure}

Therefore, for a given batch $\mathbf{X}= \{X_1,X_2,\ldots,X_B\}$, we first calculate the average cosine similarity $S_B$ between these samples. Then, the running threshold $S$ is accumulated with the feature cosine similarity across iterations, representing the overall status of feature condensation. If the running threshold is higher than the pre-defined constant FCT $\tau$, features are ``condensed,'' which requires BN to alleviate such similarity with running statistics. In contrast, when the running average feature similarity $S$ is lower than FCT $\tau$, we use statistics accumulated before and do not update statistics with the information in the current batch, to enhance learning performance. Specifically, we determine our statistics in UBN as shown in both Eq.~(\ref{eq:FCT}) and Figure~\ref{fig:ubn}.
\begin{equation}
    \label{eq:FCT}
    \begin{aligned}
       & S_B = \frac{1}{B} \sum_{i,j\in B,i\neq j}\cos(X_i,X_j)\\
       &S \gets m'S + (1-m') S_B \\
            & \mu \gets \left\{
    \begin{aligned}
      m\mathbb{E}[\mathbf{X}]& +(1-m)\mu_{\text{old}} & \ \text{if} \ S > \tau   \\
      & \mu_{\text{old}} & \ \text{if} \  S \leq \tau
    \end{aligned}
    \right. \\
    & \sigma^2 \gets \left\{
    \begin{aligned}
    m\text{Var}[\mathbf{X}]& +(1-m)\sigma^2_{\text{old}} & \ \text{if} \  S > \tau 
    \\
    & \sigma^2_{\text{old}} & \ \text{if} \ S \leq \tau 
    \end{aligned}
    \right.
    \end{aligned}
\end{equation}
where $m'$ denotes the momentum used for feature condensation threshold $\tau$, $m$ denotes the momentum used for mean and variance in normalization.



\begin{algorithm}[tb!]
\caption{Unified Batch Normalization}
\begin{algorithmic}[1]
    \State \textbf{Input:} a batch of features $\mathbf{X}$ with size $B \times C \times W \times H$
    \State \textbf{Parameters:} momentum for normalization $m$, feature condensation threshold (FCT) $\tau$, momentum for FCT $m'$
    \State \textbf{Initialize:} Running mean $\mu \gets 0$, Running variance $\sigma^2 \gets 0$, Feature condensation score $S \gets 0$
    \State \textbf{Stage 1: Determining statistics $\mu$ and $\sigma$}
    \State \quad $S_B = \frac{1}{B} \sum_{i,j\in B,i\neq j}\cos(X_i,X_j)$,\quad $S \gets m'S + (1-m') S_B$
    \State \quad $ \mu \gets \left\{
    \begin{aligned}
      m\mathbb{E}[\mathbf{X}]& +(1-m)\mu_{\text{old}} & \ \text{if} \ S > \tau   \\
      & \mu_{\text{old}} & \ \text{if} \  S \leq \tau
    \end{aligned}
    \right. $
    \State \quad $ \sigma^2 \gets \left\{
    \begin{aligned}
    m\text{Var}[\mathbf{X}]& +(1-m)\sigma^2_{\text{old}} & \ \text{if} \  S > \tau 
    \\
    & \sigma^2_{\text{old}} & \ \text{if} \ S \leq \tau
    \end{aligned}
    \right. $
    \State \textbf{Stage 2: Normalization with given statistics}
    \State \quad Centering Rectification: $\mathbf{X'} = \mathbf{X} + w_c \odot \mathbf{K(\mathbf{X})}$
    \State \quad  Centering: $\mathbf{\bar{X}'} = \mathbf{X'} - \mu$
    \State \quad Scaling: $\mathbf{\hat{X}'} = \frac{\mathbf{\bar{X}}'}{\sqrt{\sigma^2 + \varepsilon}}$
    \State \quad Scaling Rectification: $\mathbf{\hat{X}''} = \mathbf{\hat{X}'}\cdot \text{sigmoid}(w_s \odot \mathbf{K(\mathbf{\hat{X}'})} + b_s)$
    \State \quad  Affine Rectification: $\mathbf{Y} = \text{conv}(\mathbf{\hat{X}''})$
\end{algorithmic}
\label{alg:ubn}
\end{algorithm}
\vspace{-10pt}

\subsubsection{Normalization Rectification}
Although directly applying FCT on most DNNs gets performance gain by stabilizing the running statistics in each batch, in certain scenarios, the conventional application of BN may still face challenges due to the misalignment between each instance in the mini-batch. This inconsistency can weaken the performance of BN in real-world applications. Therefore, we aim to rectify these operations using instance statistics during normalization, enhancing the model's ability to capture instance-specific representations.

The rectification process involves additional trainable weights by adopting the idea from RBN \cite{gao2021representative}. Specifically, during the centering operation, we learn weights $w_c$ to capture the importance of representative features. We apply average pooling on the given input feature and use $w_c$ to obtain the residual rectification of our input features. During the scaling operation, we learn weights $w_s$ and $b_s$ to capture the intensity of de-centered features to maintain better feature intensity. We also apply average pooling on the given decentralized features and use $w_s$ and $b_s$ to obtain the rectified intensity scaling factor. Specifically, the normalization rectification is shown as follows:
\begin{equation}
    \label{eq:rbn}
    \begin{aligned}
        & \text{Centering Rectification:} \ \mathbf{X'} = \mathbf{X} + w_c \odot \mathbf{K(\mathbf{X})}, \\
        & \text{Centering:} \ \mathbf{\bar{X}'} = \mathbf{X'} - \mu, \\
        & \text{Scaling:} \ \mathbf{\hat{X}'} = \frac{\mathbf{\bar{X}}'}{\sqrt{\sigma^2 + \varepsilon}}, \\
        & \text{Scaling Rectification:} \
         \mathbf{\hat{X}''} = \mathbf{\hat{X}'}\cdot \text{sigmoid}(w_s \odot \mathbf{K(\mathbf{\hat{X}'})} + b_s), \\
    \end{aligned}
\end{equation}
where $w_c,w_s,b_s$ are learned weights and bias, and $\mathbf{K}$ is the average pooling operation.
\vspace{-10pt}

\subsubsection{Affine Rectification}
After rectifying the normalization operation in BN, we aim to tackle the limitations with the traditional affine transformation within the BN. The conventional affine operation applies global scaling and shifting to the normalized input, potentially overlooking intricate local patterns. Motivated to capture fine-grained details and enhance performance, we propose an affine rectification strategy that leverages convolutional operations to introduce local information into the affine transformation. Inspired by BNET \cite{10012548}, we propose an affine rectification as shown in Eq.~(\ref{eq:bnet}).
\begin{equation}
    \label{eq:bnet}
    \begin{aligned}
        & \text{Affine Rectification:} \ \mathbf{Y} = \text{conv}(\mathbf{\hat{X}''}) \\
    \end{aligned}
\end{equation}
where $\mathbf{Y}$ is obtained using convolutional operation on the channel of features $\mathbf{X}$.

\section{Experiments}
\label{sec:exp}
We evaluate UBN in image classification on the ImageNet \cite{russakovsky2015imagenet}, CIFAR-100 \cite{krizhevsky2009learning}, and CIFAR-10 \cite{krizhevsky2009learning} datasets. We also generalize UBN to Object Detection and Instance Segmentation on the COCO \cite{lin2015microsoft} dataset. To validate the design of UBN, we further perform ablation studies on each rectification in each component of BN. Note that UBN is not limited to these evaluations and can benefit a general class of architectures.

\begin{table}[tb!]
  \caption{Classification performance of different normalization methods on the ImageNet.}
  \vspace{-10pt}
  \label{table:imagenet}
  \centering
  \begin{tabular}{@{}lcccc@{}}
    \toprule
    \textbf{ImageNet} & \makecell{Normalization} & \makecell{Top-1 Accuracy (\%)} & \makecell{Top-5 Accuracy (\%)} \\
    \midrule
    \multirow{2}{*}{ResNet-50}  & BN & 70.70 & 89.94 \\
                                & UBN($\tau=0.15$) & \textbf{73.97} & \textbf{91.66} \\
    \midrule
    \multirow{2}{*}{ResNet-101} & BN  & 68.58  & 88.26  \\
                                & UBN($\tau=0.15$)  & \textbf{72.90}  & \textbf{91.02}  \\
    \midrule
    \multirow{2}{*}{ResNeXt-50} & BN  & 69.33  & 88.64  \\
                                & UBN($\tau=0.15$)  &\textbf{72.19}   & \textbf{90.62}  \\
    \midrule
    \multirow{2}{*}{Res2Net-50} & BN  & 69.45  & 88.93  \\
                                & UBN($\tau=0.15$)  & \textbf{72.73}  & \textbf{90.88}  \\
    \bottomrule
  \end{tabular}
  \vspace{-10pt}
\end{table}

\subsection{Implementation Details}
This section delineates the implementation specifics for various models and datasets within our image classification experiments. In this section, we report our implementation details for each model and each dataset. In our proposed UBN, we incorporate RBN \cite{gao2021representative} for centering and scaling rectification by default and BNET \cite{10012548} as our default affine rectification method with its kernel size in the convolutional layer set to 3. We compute the feature cosine similarity between samples for each batch during each training iteration and evaluate the feature condensation at the end of each epoch. We have set the default FCT $\tau$ at 0.15, the hyper-parameter chosen without necessitating further specification. We use ``test accuracy'' by default to refer to ``top-1 test accuracy.''

\subsubsection{On ImageNet} Consistent with common practices, we applied data augmentation and randomly cropped images to $224\times 224$ pixels for training ResNets \cite{he2015deep}, ResNeXt-50 \cite{xie2017aggregated}, and Res2Net-50 \cite{Gao_2021}. ResNet-50 was trained using four Nvidia A100 GPUs and the other networks were trained using eight Nvidia A100 GPUs. Each model was trained for 90 epochs. We adopted a multi-step decay process for the learning rate. The optimizer was SGD, with an initial learning rate of 0.1, a momentum of 0.9, and a weight decay of 1e-4. The learning rate was programmed to decrease by 0.1 at the 30th and 60th epochs. The batch size per GPU was set to 256. In each model, the BN layer was substituted with our as UBN, detailed in Table~\ref{table:imagenet}.

\subsubsection{On CIFAR-100} We also conducted experiments on the CIFAR-100 using three Nvidia 3090 GPUs. Our process followed typical data augmentation practices, including random cropping of images to $32\time 32$ pixels for training ResNets \cite{he2015deep}, ResNeXts \cite{xie2017aggregated}, VGGs \cite{simonyan2015deep}, and Inception-v4s \cite{szegedy2016inceptionv4}. Each model underwent 200 training epochs. We introduced a warm-up phase in the first epoch, where the learning rate increased linearly with each batch to stabilize early training. After this, we employed a multi-step decay for the learning rate. Using SGD as the optimizer, we started with an initial learning rate of 0.1, a momentum of 0.9, and a weight decay of 5e-4. The learning rate was set to decrease by 0.2 at the 60th, 120th, and 160th epochs. We used a batch size of 128 per GPU. The BN layer in each model was replaced with our UBN, as shown in Table~\ref{table:cifar100}.

\begin{table}[tb!]
  \caption{Classification performance of different normalization methods on the CIFAR-100.}
  \vspace{-10pt}
  \label{table:cifar100}
  \centering
  \begin{tabular}{@{}lcccc@{}}
    \toprule
    \textbf{CIFAR-100} & \makecell{Normalization} & \makecell{Top-1 Accuracy (\%)} & \makecell{Top-5 Accuracy (\%)} \\
    \midrule
    \multirow{7}{*}{ResNet-50}  & IN & 77.68 & 94.20 \\
                                & GN & 51.71 & 78.55 \\
                                & BN & 77.02 & 93.87 \\
                                & IEBN & 78.13 & 94.39 \\
                                & RBN & 77.79 & 94.28 \\
                                & BNET & 77.68 & 94.16 \\
                                & UBN($\tau=0.15$) & \textbf{78.67} & \textbf{94.40} \\
    \midrule
    \multirow{2}{*}{ResNet-101} & BN  & 77.18  & 93.87  \\
                                & UBN($\tau=0.1$)  & \textbf{78.48}  & \textbf{94.28}  \\
    \midrule
    \multirow{2}{*}{ResNet-152} & BN  & 77.87  & \textbf{94.40}  \\
                                & UBN($\tau=0.15$)  & \textbf{78.28}  & 94.38  \\
    \midrule
    \multirow{2}{*}{ResNeXt-50} & BN  & 77.49  & 93.69  \\
                                & UBN($\tau=0.15$)  & \textbf{77.81}  & \textbf{94.46}  \\

    \midrule
    \multirow{2}{*}{VGG-13} & BN  & 71.03  & 90.53  \\
                                & UBN($\tau=0.25$)  & \textbf{71.37}  & \textbf{91.06}  \\
    \midrule
    \multirow{2}{*}{Inception v4} & BN  & 75.60  & 92.83  \\
                               & UBN($\tau=0.25$)  & \textbf{76.51}  & \textbf{93.53}  \\
    \bottomrule
  \end{tabular}
\vspace{-10pt}
\end{table}

\subsubsection{On CIFAR-10} Our experiments were carried out on the CIFAR-10 dataset using three Nvidia 3090 GPUs. We adhered to standard data augmentation techniques, including random cropping of images to $32\time 32$ pixels for training ResNets \cite{he2015deep} and VGGs \cite{simonyan2015deep}. Each model was trained for a total of 200 epochs. For learning rate management, we applied a multi-step decay strategy. Our optimization choice was SGD, starting with a learning rate of 0.1, a momentum of 0.9, and a weight decay rate of 1e-4. We programmed the learning rate to reduce by 0.1 at the 100th and 150th epochs. Each GPU was assigned a batch size of 128. In each model, the BN layer was replaced with our proposed UBN, as detailed in Table~\ref{table:cifar10}.

\begin{figure}[tb!]
    \centering
    \includegraphics[width=0.9\linewidth]{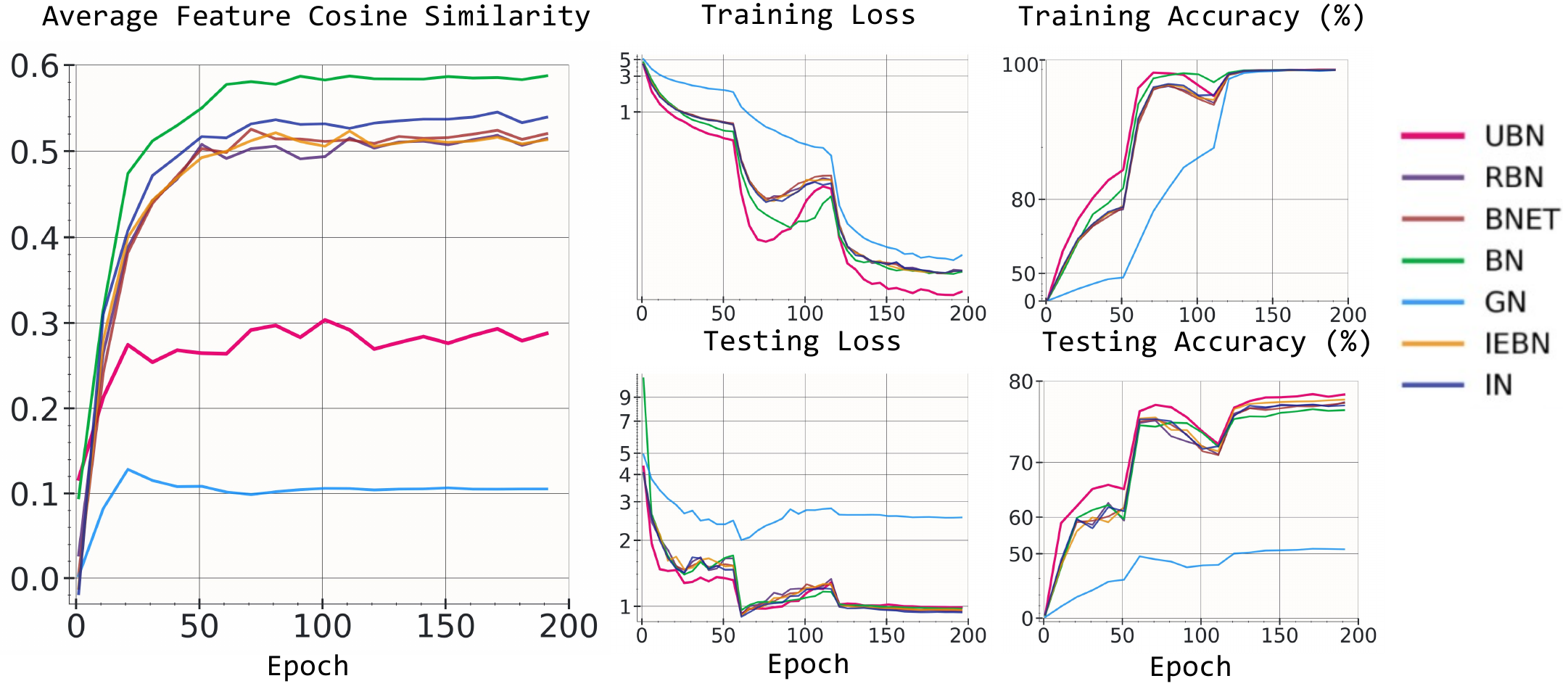}
    \captionsetup{skip=3pt}
    \caption{The learning curves and feature condensation curves of UBN and other normalization methods with ResNet-50 on the CIFAR-100 dataset. UBN can significantly reduce the feature condensation phenomenon without abandoning the batch statistics like GN \cite{wu2018group}, LN \cite{ba2016layer}, and IN \cite{ulyanov2017instance}. The leveraging of feature condensation threshold boost the stability of training, and the unification of rectifications of each component in BN boost the performance of UBN compared with existing normalization methods. We used non-uniform axis scale to visualize the differences between UBN and other SOTA normalization methods in both loss and accuracy.
    }
    \label{fig:baselines}
\end{figure}

\begin{table}[tb!]
  \caption{Classification performance of different normalization methods on the CIFAR-10.}
  \vspace{-10pt}
  \label{table:cifar10}
  \centering
  \begin{tabular}{@{}lcccc@{}}
    \toprule
    \textbf{CIFAR-10} & \makecell{Normalization} & \makecell{Top-1 Accuracy (\%)} & \makecell{Top-5 Accuracy (\%)} \\
    \midrule
    \multirow{2}{*}{ResNet-18}  & BN & 93.65 & \textbf{99.85} \\
                                & UBN($\tau=0.025$) & \textbf{93.80} & 99.81 \\
    \midrule
    \multirow{2}{*}{ResNet-34}  & BN & 93.02 & 99.76 \\
                                & UBN($\tau=0.15$) & \textbf{93.98} & \textbf{99.83} \\
    \midrule
    \multirow{2}{*}{ResNet-50} & BN  & 92.26  & 99.84  \\
                                & UBN($\tau=0.15$)  & \textbf{94.32}  & \textbf{99.85}  \\
    \midrule
    \multirow{2}{*}{ResNet-101} & BN  & 93.19  & 99.87  \\
                                & UBN($\tau=0.25$)  & \textbf{94.56}  & \textbf{99.89}  \\
    \midrule

    \multirow{2}{*}{VGG-11} & BN  & 91.14  & 99.65  \\
                                & UBN($\tau=0.25$)  & \textbf{91.48}  & \textbf{99.66}  \\
             
    \bottomrule
  \end{tabular}
  \vspace{-10pt}
\end{table}

\subsection{Performance Evaluation}

This section presents our findings on image classification for vision tasks, comparing our results with SOTA normalization methods across various datasets and DNNs.

\subsubsection{Comparison of other normalization methods} 
On the CIFAR-100 dataset, we compared UBN with existing BN variants, as detailed in Table~\ref{table:cifar100} and Figure~\ref{fig:baselines}. The batch size was set to 128 for each GPU. GN \cite{wu2018group}, which doesn't utilize batch statistics, underperforms compared to standard BN in typical training setups. UBN, however, preserves the advantages of batch dependency while incorporating conditional adaptive running statistics. It also outperforms other leading normalization methods like RBN \cite{gao2021representative}, IEBN \cite{liang2019instance}, and BNET \cite{10012548}. These methods tend to focus on specific aspects of BN, making targeted adjustments but overlooking a comprehensive perspective of BN from a running statistics viewpoint. UBN consistently demonstrates higher accuracy early in training, maintaining this advantage to the end.

\subsubsection{Incorporating into different network architectures}
As a modification of the original Batch Normalization, our developed Unified Batch Normalization (UBN) is designed to replace BN layers in various networks, thereby improving classification performance across multiple datasets including ImageNet, CIFAR-100, and CIFAR-10, as indicated in Tables~\ref{table:imagenet}, \ref{table:cifar10}, and \ref{table:cifar100}. DNNs using UBN, particularly ResNets, show superior top-1 accuracies over those employing standard BN. On ImageNet, notable enhancements are observed in accuracies with ResNet-50 gaining a 3.27\% improvement, ResNet-101 showing a 4.32\% increase, and ResNext-50 with a 2.86\% improvement. Res2Net-50 also achieving a 3.28\% gain. On CIFAR-10 and CIFAR-100, UBN based ResNets exhibit the most notable improvement over BN based ResNets, achieving performance gains of 2.1\% and 1.4\% respectively. For VGG architectures, the implementation of UBN results in a 0.3\% accuracy increase on both datasets, compared to BN based VGGs.

\subsubsection{Generalization to Typical Vision Tasks}
Our proposed UBN method is designed to enhance the representational capabilities of models across a variety of tasks, effectively serving as a superior alternative to traditional BN. We evaluated UBN's performance on typical vision tasks by integrating it into Faster-RCNN for object detection and Mask-RCNN for instance segmentation within the COCO dataset. Each model underwent training for a duration of 12 epochs, utilizing a multi-step decay strategy for optimal learning rate adjustment. The optimization was carried out using SGD with an initial learning rate set at 0.02, momentum at 0.9, and weight decay at 1e-4. We used a batch size of 2 for each GPU (8 in total) during the training process. The empirical results, as detailed in Table \ref{table:obj-det} and Table \ref{table:ins-seg}, demonstrate that UBN substantially surpasses BN, achieving an improvement exceeding 4\% in mean Average Precision (AP).

\begin{figure}[tb!]
    \centering
    \includegraphics[width=0.99\linewidth]{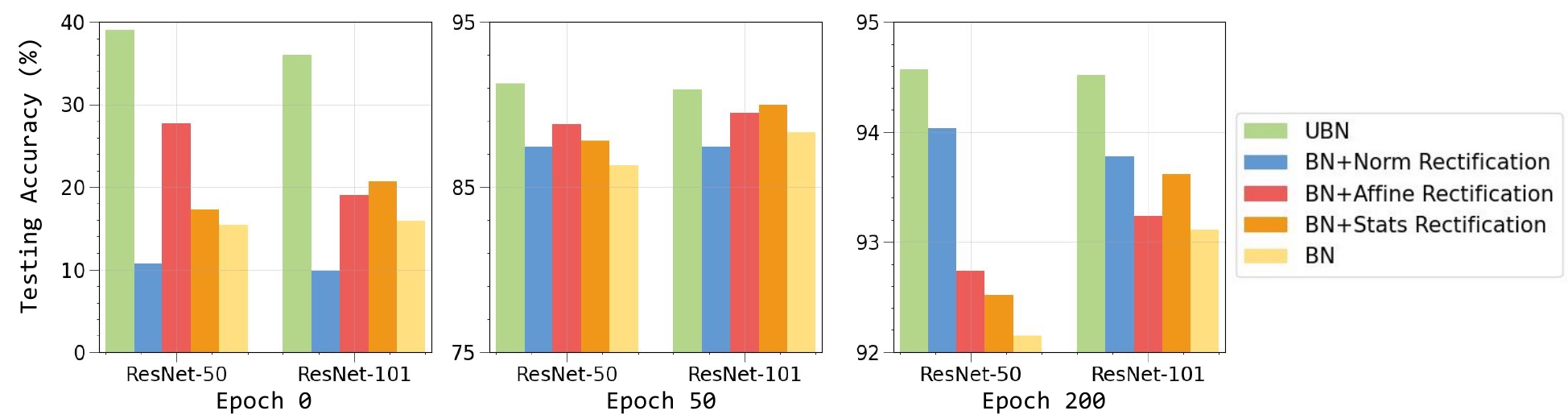}
    \captionsetup{skip=3pt}
    \caption{Effectiveness ablation (testing accuracies) on ResNet-50 and ResNet-101 trained on the CIFAR-10 dataset with batch size 128 for 200 epochs. We plot the accuracies at the end of Epoch 0, 50, and 200. Each rectification boosts the testing performance of BN.
    }
    \label{fig:ablation-rec}
\vspace{-10pt}
\end{figure}

\begin{table}[tb!]
\caption{Object Detection Results on COCO using Faster-RCNN}
\vspace{-10pt}
  \label{table:obj-det}
\centering
    \begin{tabular}{l|cccccc}
    \toprule
     & AP & AP$_{50}$ & AP$_{75}$ & AP$_S$ & AP$_M$ & AP$_L$ \\
    \hline
    ResNet-50-BN  & 33.1 & 53.2 & 35.6 & 18.4 & 35.8 & 43.6 \\
    ResNet-50-UBN ({$\tau=0.15$}) & \textbf{37.7} & \textbf{58.6} & \textbf{40.5} & \textbf{20.8} & \textbf{40.9} & \textbf{49.5} \\
    \hline
    ResNet-101-BN  & 31.9 & 51.4 & 34.1 & 17.6 & 34.5 & 42.2 \\
    ResNet-101-UBN ({$\tau=0.15$}) & \textbf{37.6} & \textbf{58.2} & \textbf{40.1} & \textbf{20.7} & \textbf{40.7} & \textbf{50.4} \\
    \bottomrule
    \end{tabular}
\end{table}

\begin{table}[tb!]
\caption{Instance Segmentation Results on COCO using Mask-RCNN}
\vspace{-10pt}
\label{table:ins-seg}
\centering
    \begin{tabular}{l|cccccc}
    \toprule
    Box & AP & AP$_{50}$ & AP$_{75}$ & AP$_S$ & AP$_M$ & AP$_L$ \\
    \hline
    ResNet-50-BN  & 34.0 & 54.0 & 36.6 & 18.5 & 36.9 & 44.5 \\
    ResNet-50-UBN ({$\tau=0.15$}) & \textbf{38.5} & \textbf{59.3} & \textbf{41.6} & \textbf{21.3} & \textbf{41.9} & \textbf{50.5} \\
    \hline
    ResNet-101-BN  & 33.1 & 52.5 & 35.7 & 17.8 & 35.8 & 44.0 \\
    ResNet-101-UBN ({$\tau=0.15$}) & \textbf{38.6} & \textbf{59.1} & \textbf{41.4} & \textbf{22.3} & \textbf{41.6} & \textbf{51.8} \\
    
    \hline
    Mask & AP & AP$_{50}$ & AP$_{75}$ & AP$_S$ & AP$_M$ & AP$_L$ \\
    \hline
    ResNet-50-BN  & 31.1 & 50.8 & 33.0 & 13.5 & 33.1 & 46.4 \\
    ResNet-50-UBN ({$\tau=0.15$}) & \textbf{34.8} & \textbf{56.1} & \textbf{36.8} & \textbf{15.6} & \textbf{37.3} & \textbf{50.7} \\
    \hline
    
    ResNet-101-BN  & 30.3 & 49.4 & 32.0 & 12.7 & 31.8 & 45.7 \\
    ResNet-101-UBN ({$\tau=0.15$}) & \textbf{34.6} & \textbf{55.8} & \textbf{36.6} & \textbf{16.5} & \textbf{36.6} & \textbf{51.2} \\ \bottomrule
    \end{tabular}
\end{table}

\subsection{Ablation Study}
We conduct ablation experiments on our proposed UBN. The effectiveness of each rectification of the component in UBN is verified. We also investigate the effects of different batch sizes and feature condensation thresholds.

\subsubsection{Effects of each rectification in UBN} 
Our ablation studies focus on each rectification in UBN – normalization (including centering and scaling), affine, and running statistics rectification with feature condensation threshold. We conducted experiments on ResNet-50 and ResNet-101 on the CIFAR-10 dataset, as shown in Figure~\ref{fig:ablation-rec}. For the experiment on ResNet-50, BN+Stats Rectification used threshold $\tau=0.25$ and UBN used threshold $\tau = 0.15$. For experiments on ResNet-101, both BN+Stats Rectification and UBN used threshold $\tau = 0.25$. We find that each rectification type improves BN performance. When combined, these modifications result in a surprisingly efficient performance gain throughout training.

\subsubsection{Different feature condensation threshold $\tau$}
\begin{figure}[tb!]
    \centering
    \includegraphics[width=0.99\linewidth]{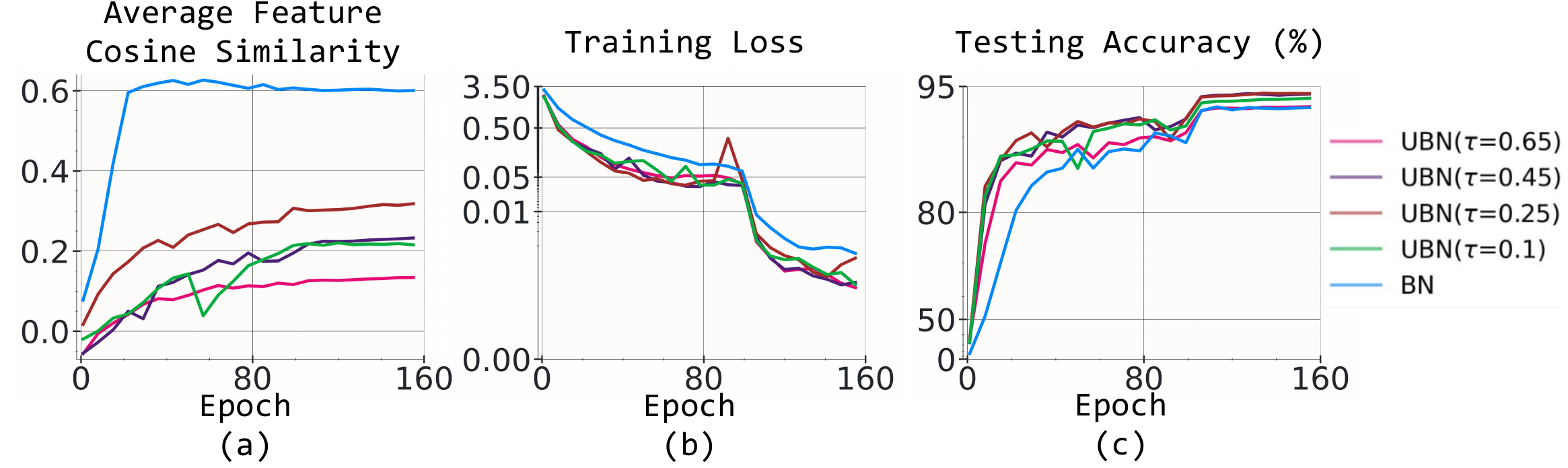}
    \captionsetup{skip=3pt}
    \caption{Ablation of testing accuracy when applying UBN with different feature condensation threshold $\tau$. (a) Average feature cosine similarity curves of the features in the last batch of each epoch before the first normalization layer (b) Training loss curves (c) Testing accuracy curves. Using the threshold can alleviate the feature condensation problem. We show the average cosine similarity for a batch of samples at the end of each epoch. We used non-uniform axis scale to visualize the differences between UBN and other normalization methods w.r.t. loss and accuracy.
    }
    \label{fig:bound}
\end{figure}

We analyze the effect of feature condensation threshold $\tau$, which determines the running statistics in UBN. We show the results of the ResNet-101 model on the CIFAR-10 dataset. The specific statistics of each batch of samples is unpredictable, thus we can only provide a constant threshold to alleviate feature condensation.  As shown in Figure~\ref{fig:bound}, UBN with different threshold $\tau$ both boost the testing performance in the learning of DNNs. We demonstrate that the thresholds can dramatically alleviate the feature condensation which causes training efficiency in the early epochs. Although there are no linear or trivial relationships between the thresholds and the testing accuracies, feature condensation can be alleviated by leveraging such a threshold. \textit{For simplicity and efficiency in our experimental setup, we empirically set the threshold $\tau=0.15$. This choice, made without manual tuning, worked effectively across the majority of our tests.}

\subsubsection{Different batch sizes} 
We examine UBN's performance across different architectures on CIFAR-10 and CIFAR-100. We conducted experiments on ResNet-18, ResNet-34, and ResNet-50, as shown in Table~\ref{table:ablation-bs}. We evaluated batch sizes of 32, 64, 128, 256, and 512. We did not observe distinct correlation between batch size and test accuracy.



\begin{table}[tb!]
\caption{Effectiveness ablation (final testing accuracies) on ResNet models trained on CIFAR-10 and CIFAR-100 datasets with different batch sizes for 200 epochs.}
  \vspace{-10pt}
\label{table:ablation-bs}
\centering
\begin{tabular}{c|c|ccccc}
\toprule
\multirow{2}{*}{Dataset} & \multirow{2}{*}{Model} & \multicolumn{5}{c}{Batch Size} \\ \cline{3-7}
&  & 32 & 64 & 128 & 256 & 512 \\ \hline
\multirow{3}{*}{CIFAR-10} & ResNet-18 & 94.45 & 94.90 & 94.32 & 93.79 & 93.38 \\
 & ResNet-34 & 95.51 & 94.73 & 94.54 & 94.34 & 93.78 \\
 & ResNet-50 & 95.39 & 94.59 & 94.58 & 94.57 & 94.13 \\
\hline
\multirow{2}{*}{CIFAR-100} & ResNet-50 & 78.02 & 78.52 & 78.99 & 78.86 & 77.21 \\
 & ResNet-101 & 77.53 & 78.27 & 79.33 & 78.03 & 77.87 \\
\bottomrule
\end{tabular}
\vspace{-10pt}
\end{table}


\section{Conclusion}
\label{sec:conclusion}
Previous studies have mainly focused on improving certain Batch Normalization (BN) aspects. In contrast, our work establishes a unified approach from a feature condensation perspective. Specifically, we present Unified Batch Normalization (UBN) framework, which involves a simple yet effective feature condensation rectification and precise adjustments to each component of BN. We rectify the feature condensation with a simple pre-defined threshold to boost the robustness of running statistics, and employ rectifications on each component of BN to improve the feature representation. Furthermore, UBN can be easily integrated into existing models. By ensuring appropriate running statistics and applying comprehensive rectifications across all normalization components, UBN can significantly improve both testing performance and training efficiency.

\textbf{Limitations:} Feature condensation occurs in different scenarios throughout the training of DNNs, making the precise determination of a feature condensation threshold challenging. While UBN significantly reduces training duration, the necessity for manual adjustment of this threshold and the computation of feature similarity may cause additional time and computational cost.


%
%
\bibliographystyle{splncs04}
\bibliography{main}

\clearpage
\appendix

\section{Exploring Feature Condensation in BN in Additional Experiments}
\label{sec:supp_phenomenon}
\begin{figure}[htbp]
\vspace{-10pt}
    \centering
    \includegraphics[width=0.9\linewidth]{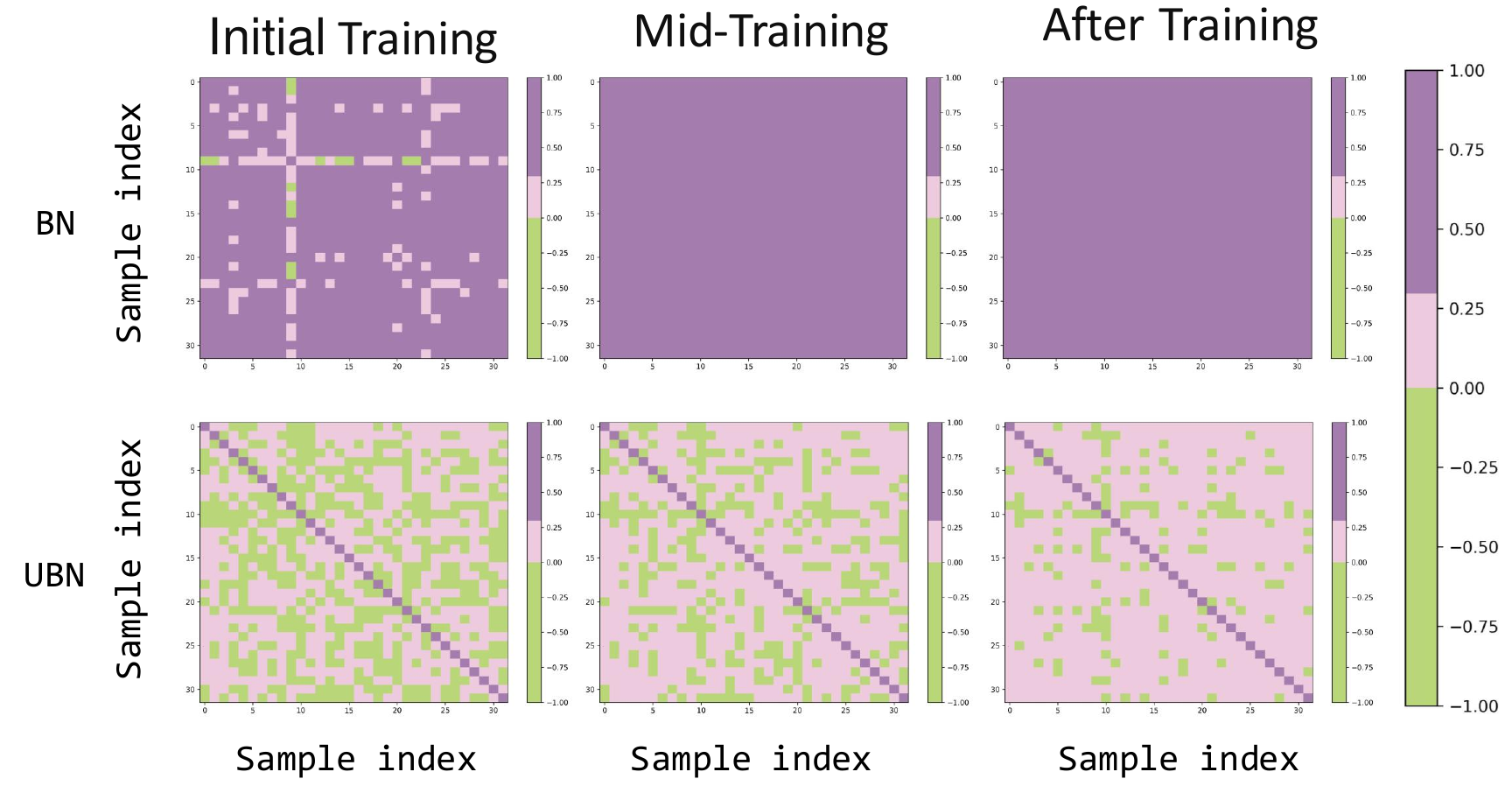}
    \captionsetup{skip=3pt}
    \caption{Feature condensation when employing BN on ResNet-50 during training on the ImageNet dataset. UBN can effectively alleviate such condensation.}
    \label{fig_supp:fct_supp_imagenet_resnet50}
\vspace{-10pt}
\end{figure}

\begin{figure}[htbp]
\vspace{-10pt}
    \centering
    \includegraphics[width=0.9\linewidth]{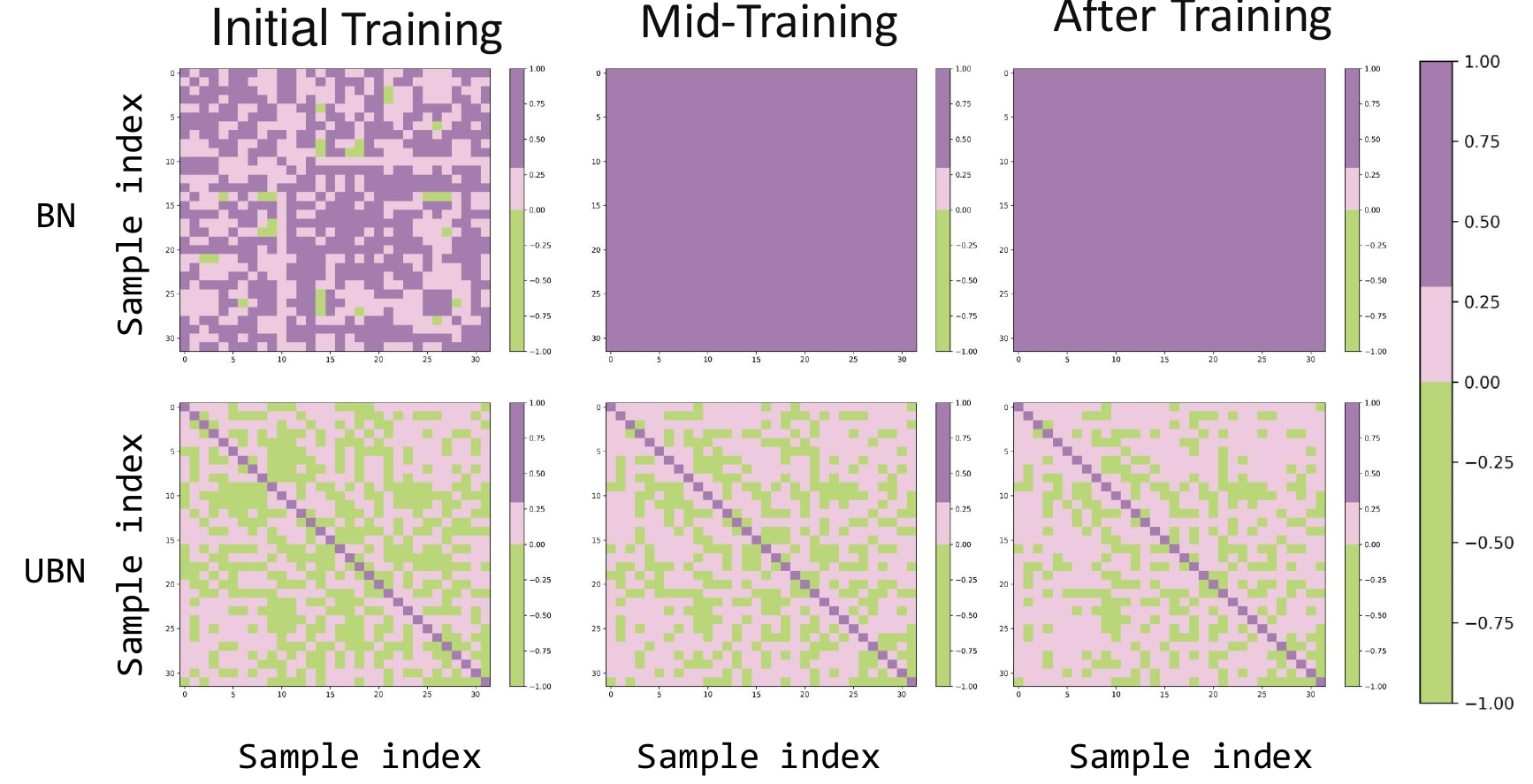}
    \captionsetup{skip=3pt}
    \caption{Feature condensation when employing BN on ResNet-101 during training on the ImageNet dataset. UBN can effectively alleviate such condensation.}
    \label{fig_supp:fct_supp_imagenet_resnet101}
\vspace{-10pt}
\end{figure}

\begin{figure}[htbp]
    \centering
    \includegraphics[width=0.9\linewidth]{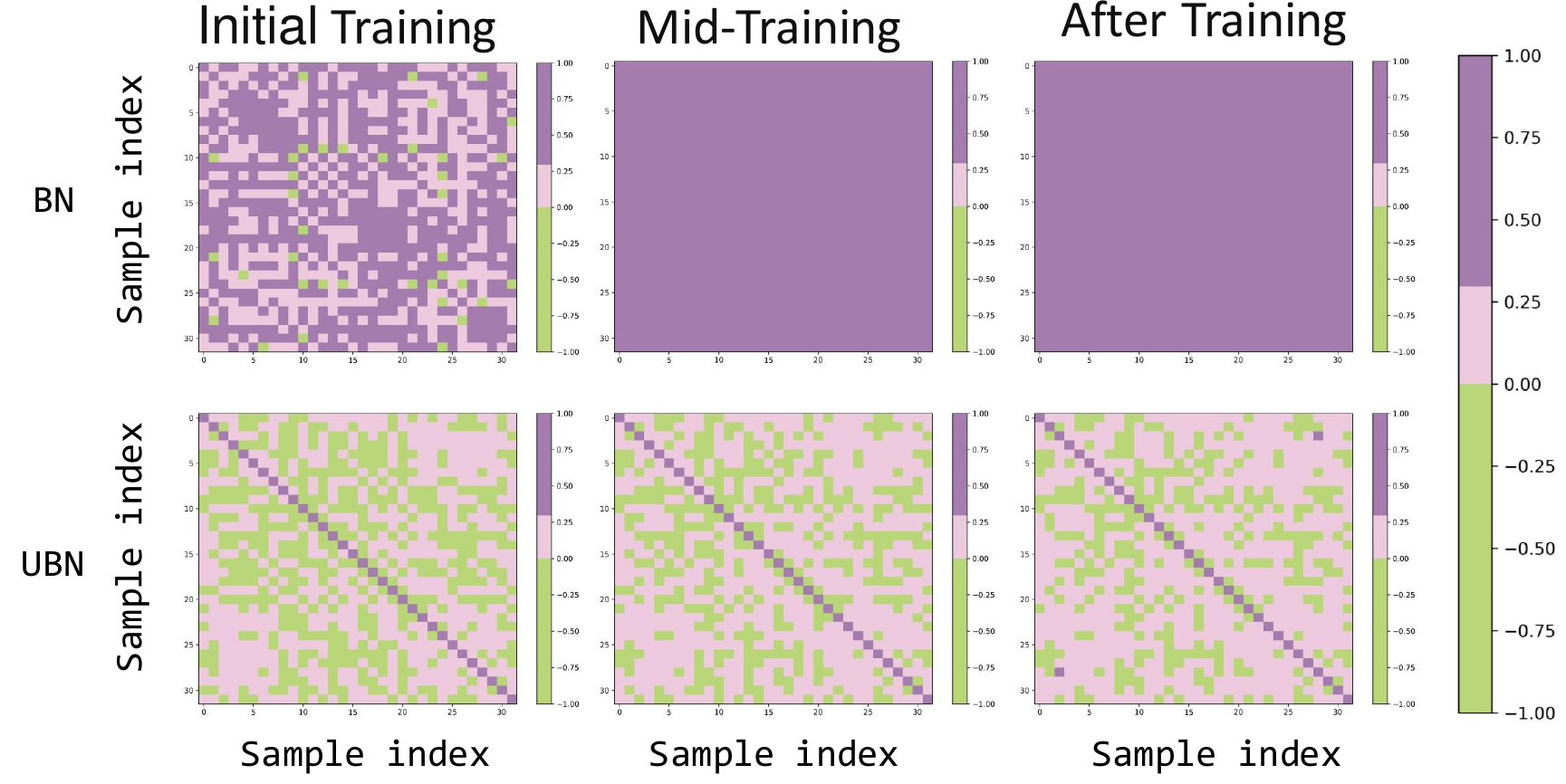}
    \captionsetup{skip=3pt}
    \caption{Feature condensation when employing BN on Res2Net-50 during training on the ImageNet dataset. UBN can effectively alleviate such condensation.}
    \label{fig_supp:fct_supp_imagenet_res2net50}
\end{figure}

\begin{figure}[htbp]
    \centering
    \includegraphics[width=0.9\linewidth]{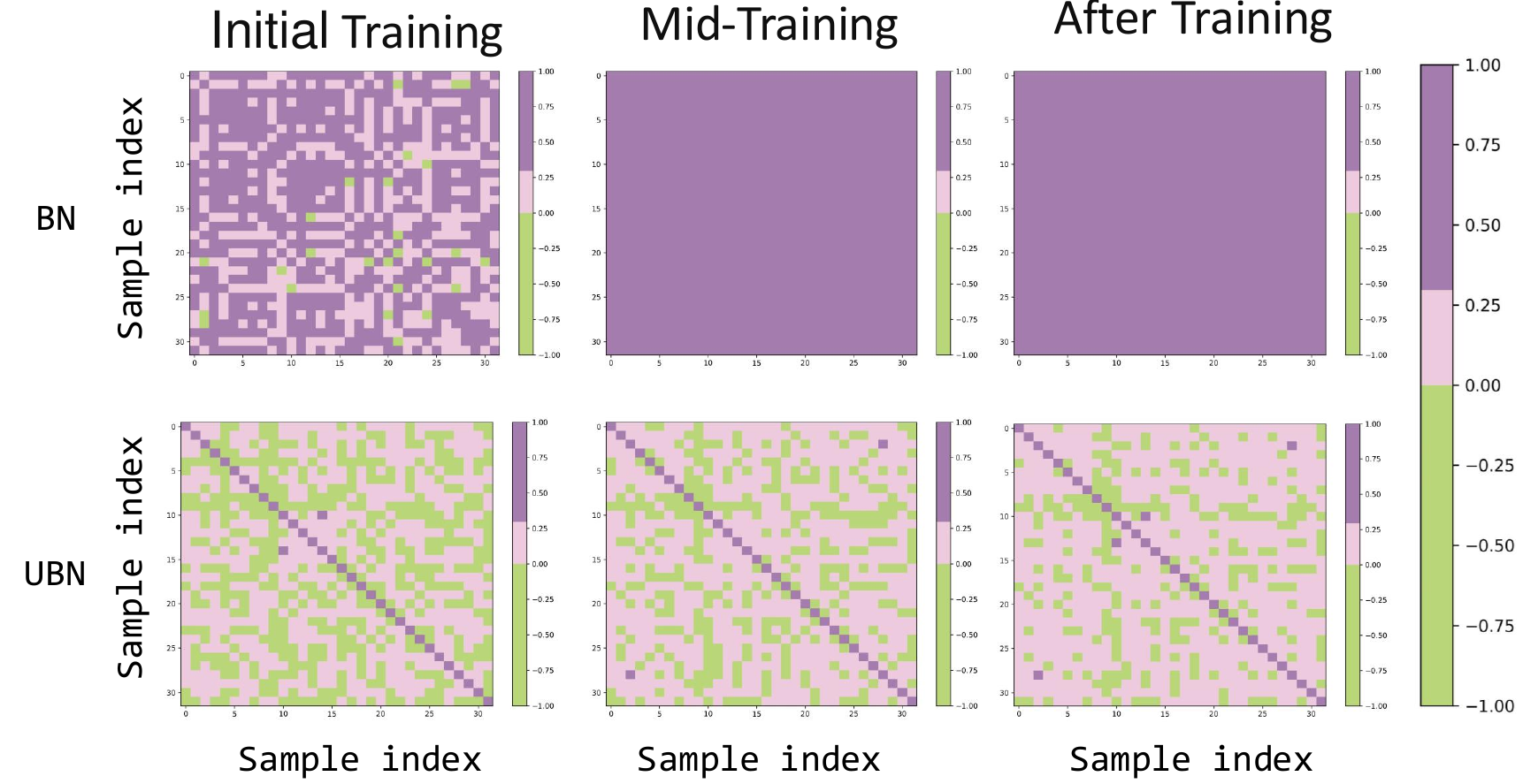}
    \captionsetup{skip=3pt}
    \caption{Feature condensation when employing BN on ResNeXt-50 during training on the ImageNet dataset. UBN can effectively alleviate such condensation.}
    \label{fig_supp:fct_supp_imagenet_resnext50}
\end{figure}

In this section, we conducted more experiments on different datasets and different models to explore the feature condensation in the training of BN, which is shown in Figure~\ref{fig:sim} before. And we aim to demonstrate that our UBN is able to alleviate such condensation with robust training statistics determination. We further explored the functionality of UBN on alleviating the feature condensation by conducting experiments on ImageNet dataset, as shown in Figure~\ref{fig_supp:fct_supp_imagenet_resnet50}, Figure~\ref{fig_supp:fct_supp_imagenet_resnet101},  Figure~\ref{fig_supp:fct_supp_imagenet_res2net50}, and Figure~\ref{fig_supp:fct_supp_imagenet_resnext50}.
We measured the cosine similarity on a fixed batch of 32 samples, which was randomly selected. We observed that UBN outperforms BN (Batch Normalization) when applied to more complex datasets. We regard features to be condensed when the similarity is over 0.3.

\clearpage
\section{Exploring Various Feature Condensation Threshold $\tau$ for UBN in Additional Experiments}
\label{sec:supp_fct}
\begin{figure}[htbp]
    \centering
    \includegraphics[width=0.7\linewidth]{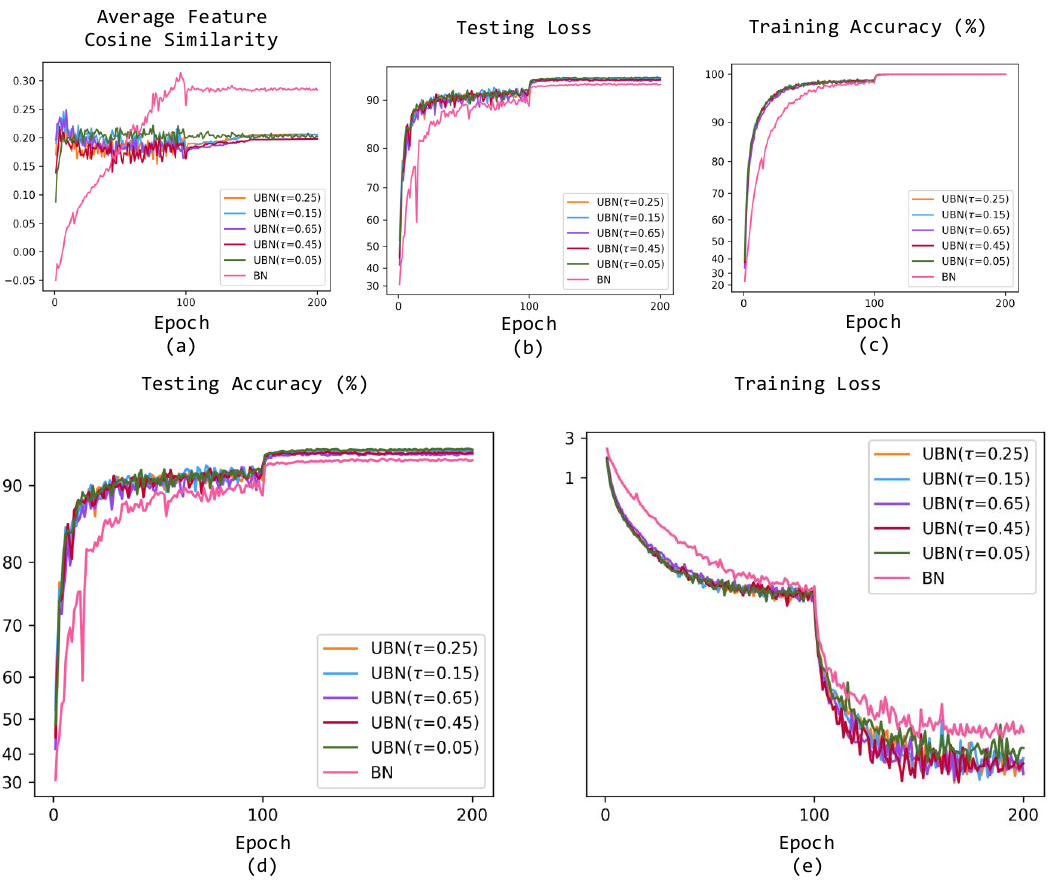}
    \captionsetup{skip=3pt}
    \caption{Feature condensation curves and learning curves of ResNet-34 on the CIFAR-10 dataset.  (a) Average feature cosine similarity of a fixed batch of samples during training. (b) Testing loss (c) Training accuracy (d) Testing accuracy (e) Training loss}
    \label{fig_supp:bound_supp_cifar10_r34}
\end{figure}

\begin{figure}[htbp]
    \centering
    \includegraphics[width=0.7\linewidth]{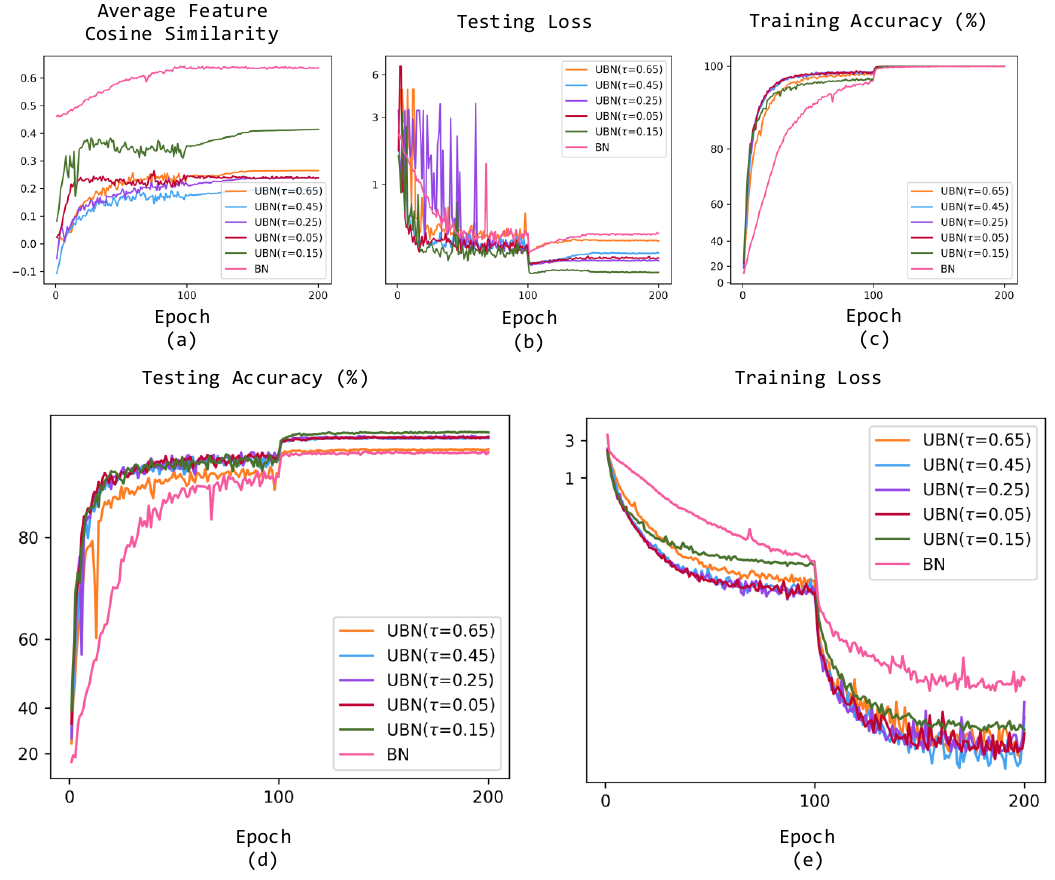}
    \captionsetup{skip=3pt}
    \caption{Feature condensation curves and learning curves of ResNet-50 on the CIFAR-10 dataset.  (a) Average feature cosine similarity of a fixed batch of samples during training. (b) Testing loss (c) Training accuracy (d) Testing accuracy (e) Training loss}
    \label{fig_supp:bound_supp_cifar10_r50}
\end{figure}

\begin{figure}[htbp]
    \centering
    \includegraphics[width=0.7\linewidth]{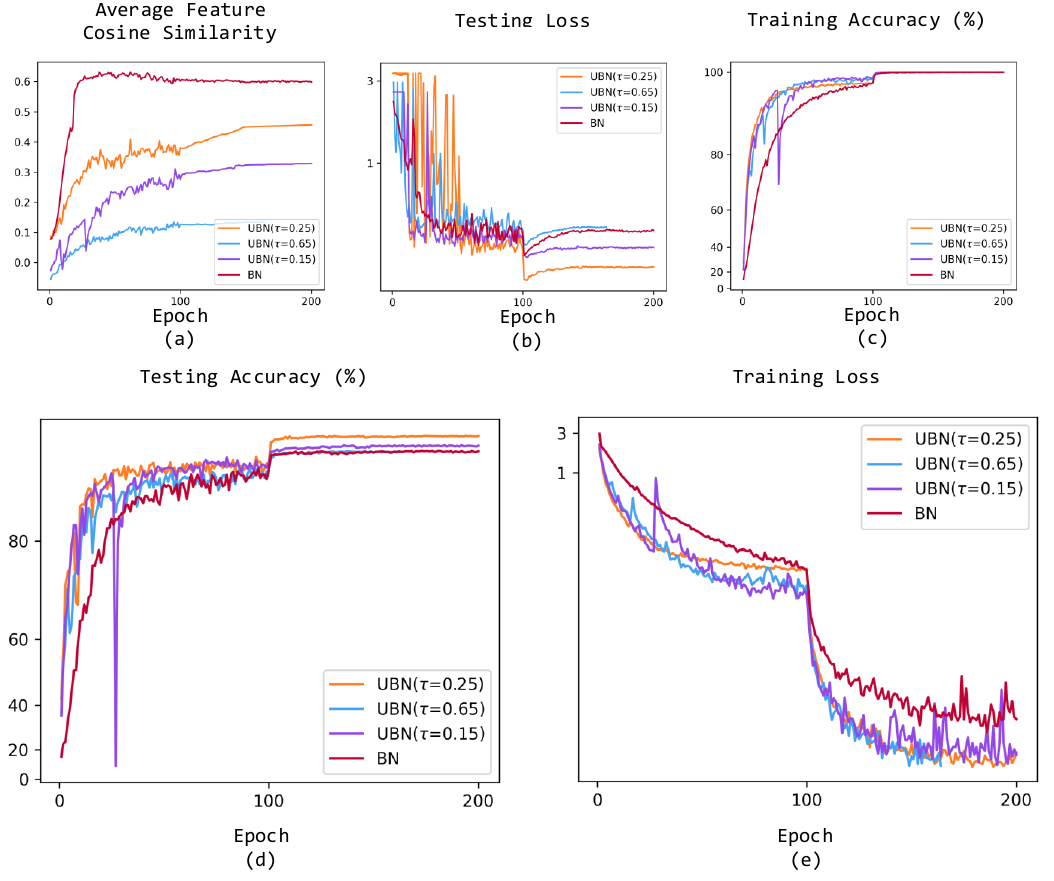}
    \captionsetup{skip=3pt}
    \caption{Feature condensation curves and learning curves of ResNet-101 on the CIFAR-10 dataset. (a) Average feature cosine similarity of a fixed batch of samples during training. (b) Testing loss (c) Training accuracy (d) Testing accuracy (e) Training loss}
    \label{fig_supp:bound_supp_cifar10_r101}
\end{figure}

\begin{figure}[htbp]
    \centering
    \includegraphics[width=0.7\linewidth]{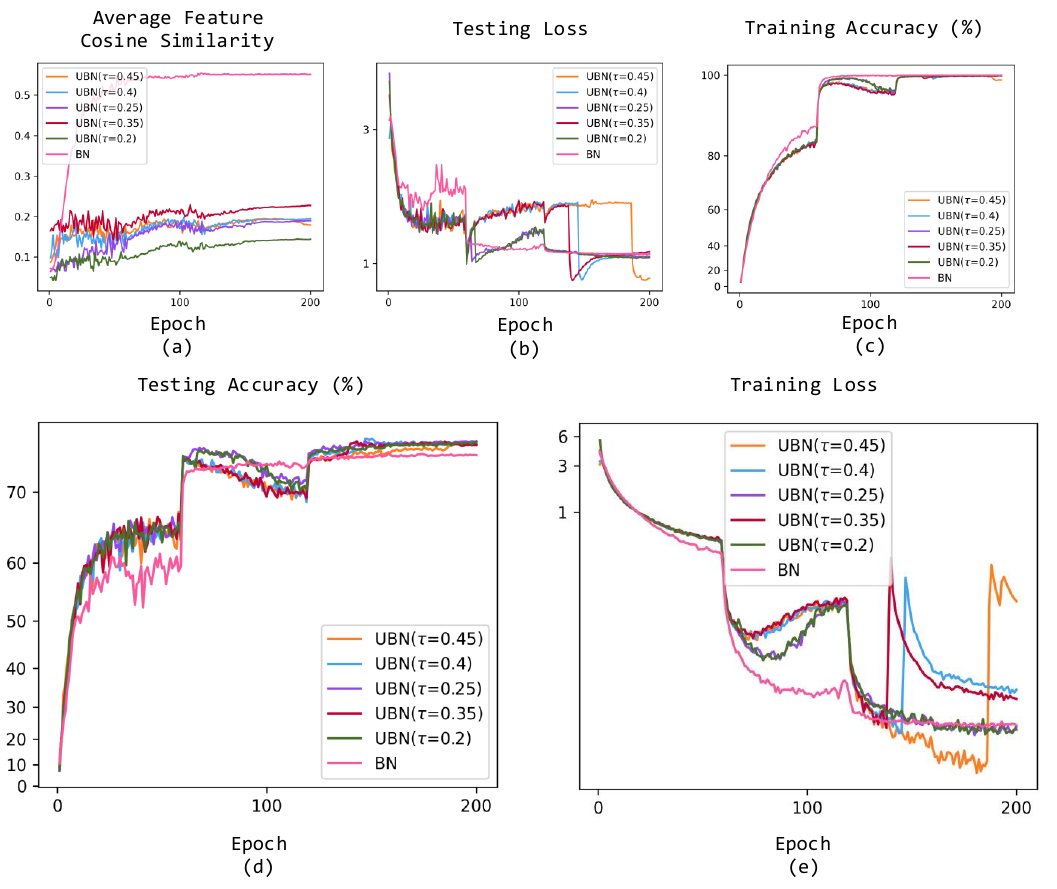}
    \captionsetup{skip=3pt}
    \caption{Feature condensation curves and learning curves of ResNet-34 on the CIFAR-100 dataset.  (a) Average feature cosine similarity of a fixed batch of samples during training. (b) Testing loss (c) Training accuracy (d) Testing accuracy (e) Training loss}
    \label{fig_supp:bound_supp_cifar100_r34}
\end{figure}

\begin{figure}[htbp]
    \centering
    \includegraphics[width=0.7\linewidth]{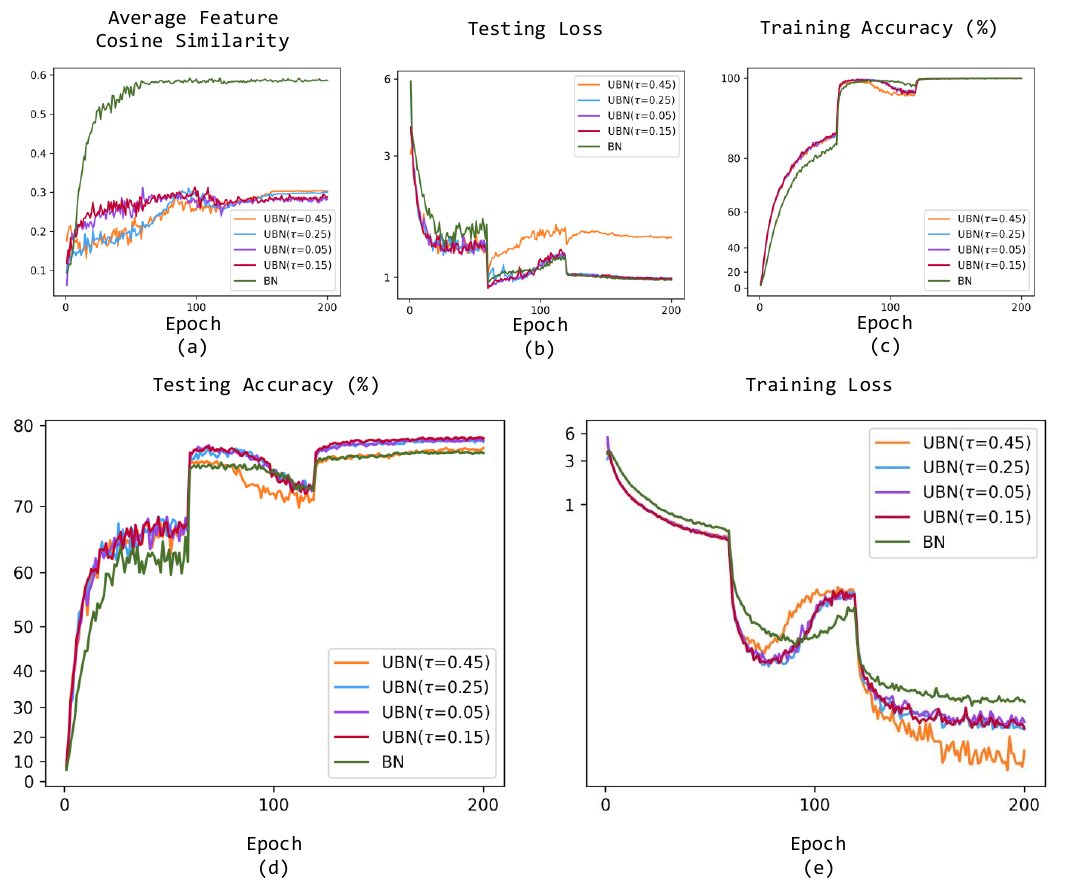}
    \captionsetup{skip=3pt}
    \caption{Feature condensation curves and learning curves of ResNet-50 on the CIFAR-100 dataset.  (a) Average feature cosine similarity of a fixed batch of samples during training. (b) Testing loss (c) Training accuracy (d) Testing accuracy (e) Training loss}
    \label{fig_supp:bound_supp_cifar100_r50}
\end{figure}

\begin{figure}[htbp]
    \centering
    \includegraphics[width=0.7\linewidth]{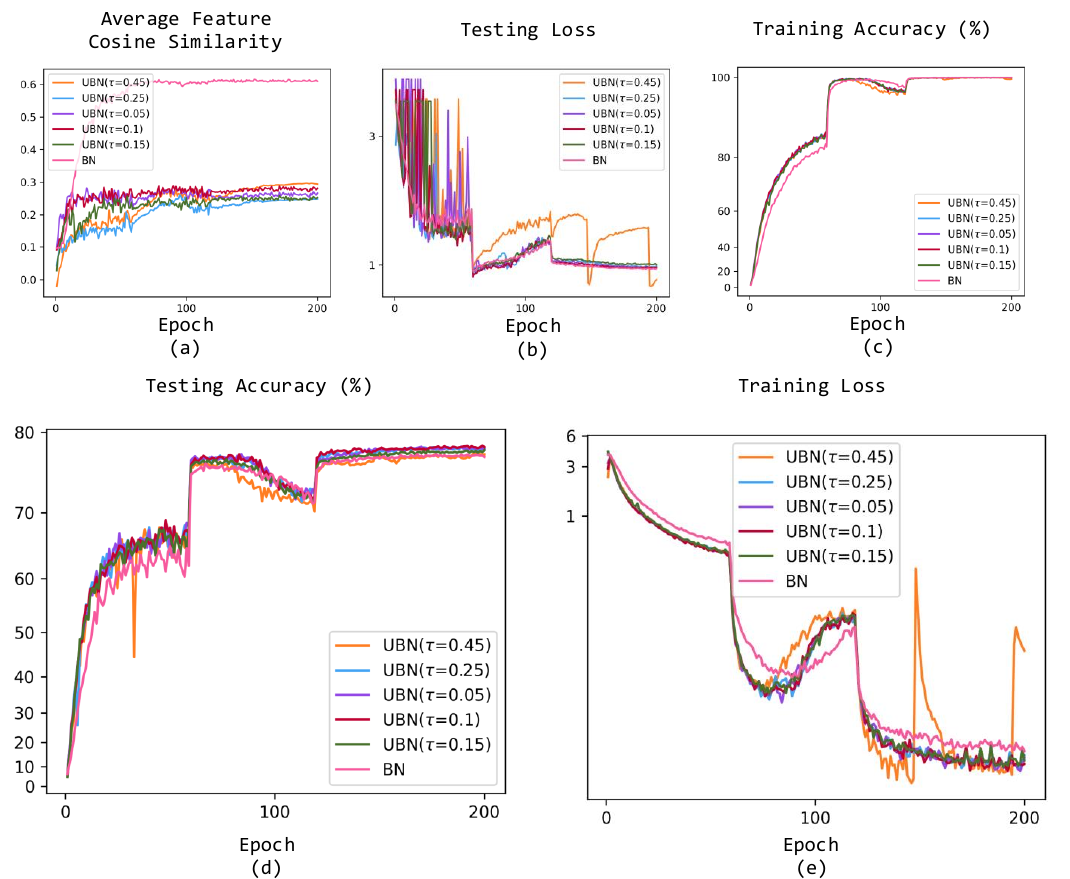}
    \captionsetup{skip=3pt}
    \caption{Feature condensation curves and learning curves of ResNet-101 on the CIFAR-100 dataset.  (a) Average feature cosine similarity of a fixed batch of samples during training. (b) Testing loss (c) Training accuracy (d) Testing accuracy (e) Training loss}
    \label{fig_supp:bound_supp_cifar100_r101}
\end{figure}

In this section, we conducted more experiments on different datasets and different models to compare different feature condensation thresholds in the training of UBN, which is shown in Figure~\ref{fig:bound} before. Extensive results on CIFAR-10 dataset are shown in Figure~\ref{fig_supp:bound_supp_cifar10_r34}, Figure~\ref{fig_supp:bound_supp_cifar10_r50}, and Figure~\ref{fig_supp:bound_supp_cifar10_r101}. Extensive results on CIFAR-100 dataset are shown in Figure~\ref{fig_supp:bound_supp_cifar100_r34}, Figure~\ref{fig_supp:bound_supp_cifar100_r50}, and Figure~\ref{fig_supp:bound_supp_cifar100_r101}.

Our experiments were conducted on the CIFAR-10 and CIFAR-100 datasets, utilizing three Nvidia 3090 GPUs. We followed the identical training settings outlined in Section ~\ref{sec:exp}. 
For CIFAR-10 dataset, each model was trained for a total of 200 epochs. For learning rate management, we applied a multi-step decay strategy. Our optimization choice is SGD, starting with a learning rate of 0.1, a momentum of 0.9, and a weight decay rate of 1e-4. We programmed the learning rate to reduce by 0.1 at the 100th and 150th epochs. Each GPU was assigned a batch size of 128.
For CIFAR-100 dataset, each model underwent 200 training epochs. We introduced a warm-up phase in the first epoch, where the learning rate increased linearly with each batch to stabilize early training. After this, we employed a multi-step decay for the learning rate. Using SGD as the optimizer, we started with an initial learning rate of 0.1, a momentum of 0.9, and a weight decay of 5e-4. The learning rate was set to decrease by 0.2 at the 60th, 120th, and 160th epochs. We maintained a batch size of 128 per GPU.


\section{Exploring the Effect of Various Batch Sizes for UBN in Additional Experiments}
\label{sec:supp_bs}
\begin{figure}[htbp]
    \vspace{-20pt}
    \centering
    \includegraphics[width=0.6\linewidth]{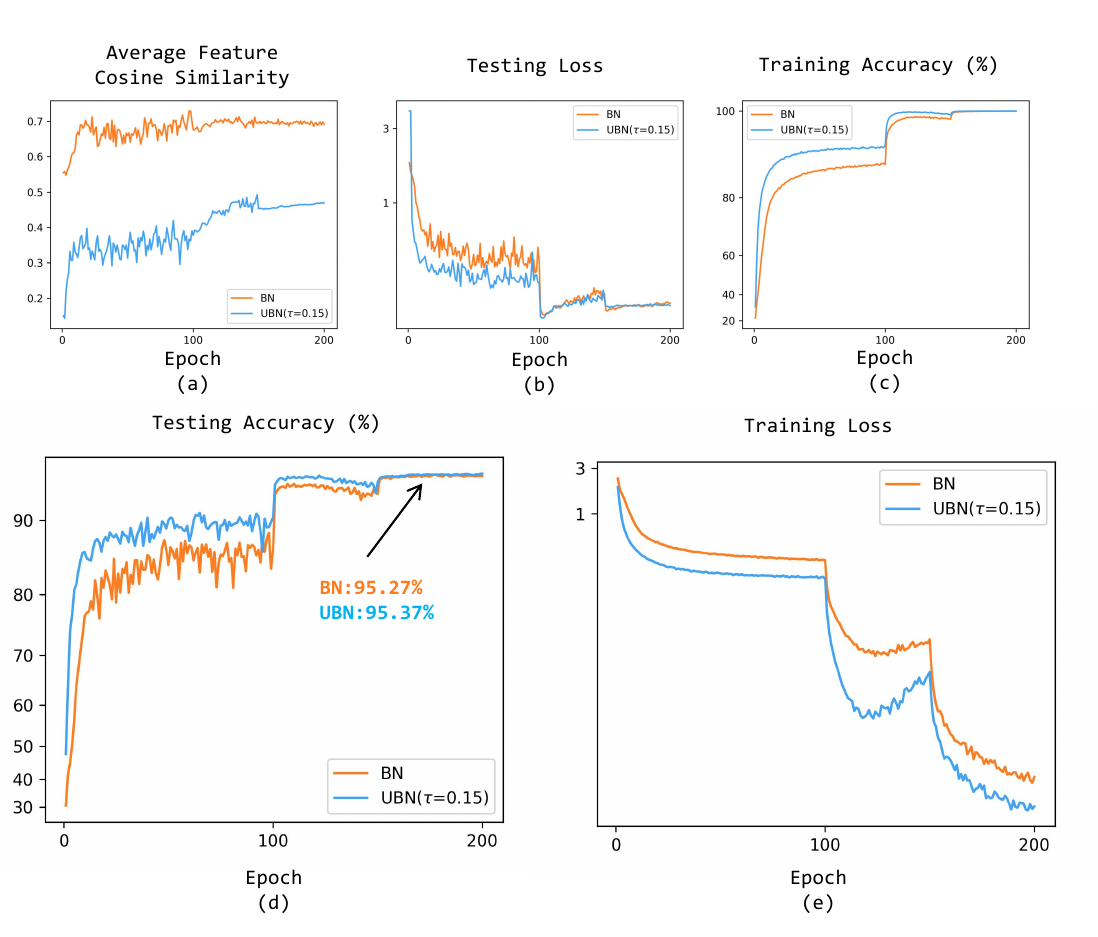}
    \captionsetup{skip=10pt}
    \caption{Feature condensation curves and learning curves of ResNet-50 on the CIFAR-10 dataset with batch size 32.  (a) Average feature cosine similarity of a fixed batch of samples during training. (b) Testing loss (c) Training accuracy (d) Testing accuracy (e) Training loss}
    \vspace{-20pt}
    \label{fig_supp:batch_size_cifar10_resnet50_bs=32}
\end{figure}

\begin{figure}[htbp]
    \centering
    \vspace{-20pt}
    \includegraphics[width=0.6\linewidth]{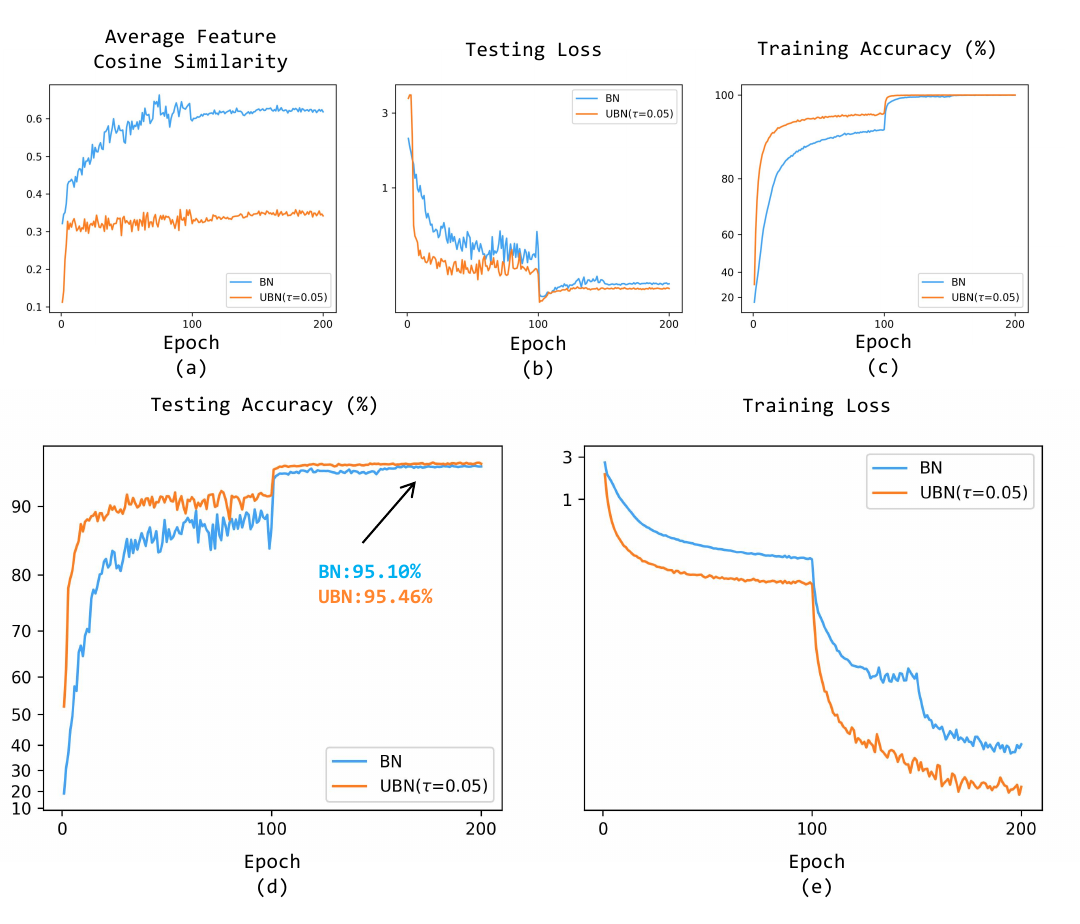}
    \captionsetup{skip=10pt}
    \caption{Feature condensation curves and learning curves of ResNet-50 on the CIFAR-10 dataset with batch size 64.  (a) Average feature cosine similarity of a fixed batch of samples during training. (b) Testing loss (c) Training accuracy (d) Testing accuracy (e) Training loss}
    \vspace{-20pt}
    \label{fig_supp:batch_size_cifar10_resnet50_bs=64}
\end{figure}

\begin{figure}[htbp]
    \centering
    \vspace{-20pt}
    \includegraphics[width=0.6\linewidth]{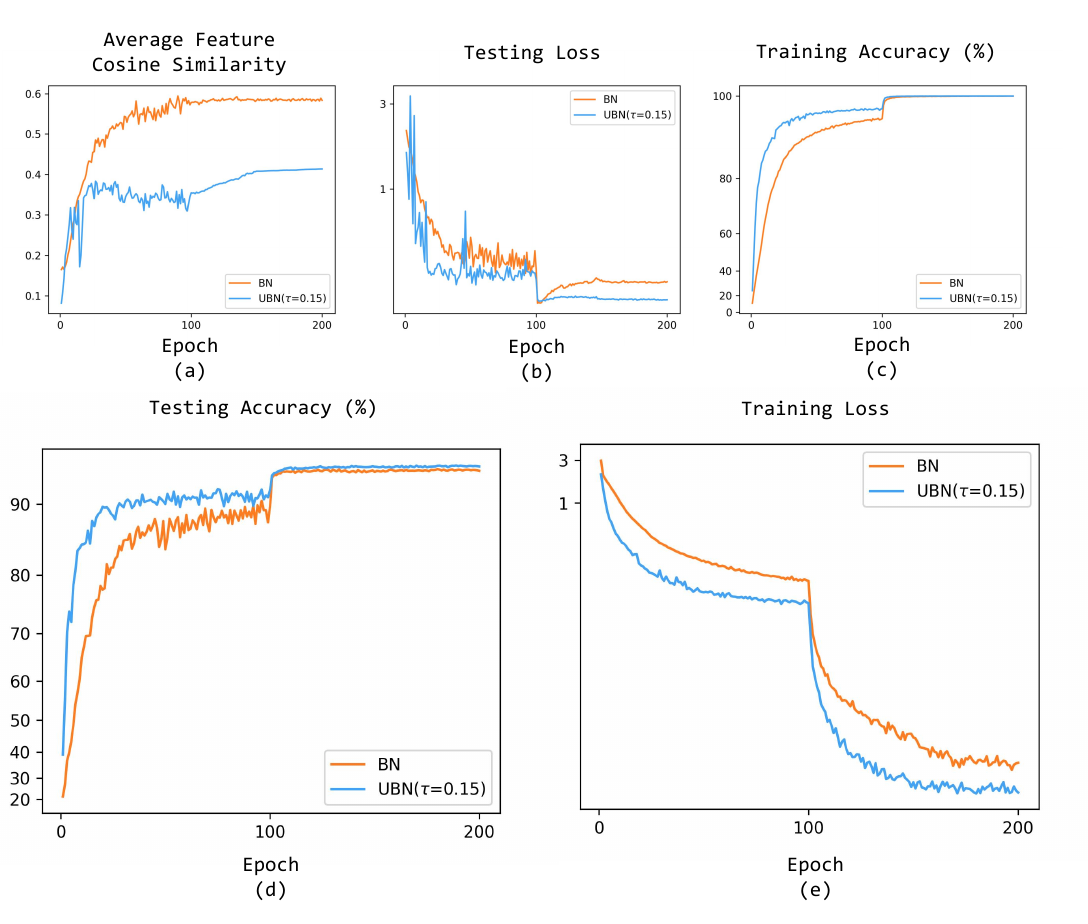}
    \captionsetup{skip=10pt}
    \caption{Feature condensation curves and learning curves of ResNet-50 on the CIFAR-10 dataset with batch size 128.  (a) Average feature cosine similarity of a fixed batch of samples during training. (b) Testing loss (c) Training accuracy (d) Testing accuracy (e) Training loss}
    \vspace{-20pt}
    \label{fig_supp:batch_size_cifar10_resnet50_bs=128}
\end{figure}

\begin{figure}[htbp]
    \centering
    \includegraphics[width=0.6\linewidth]{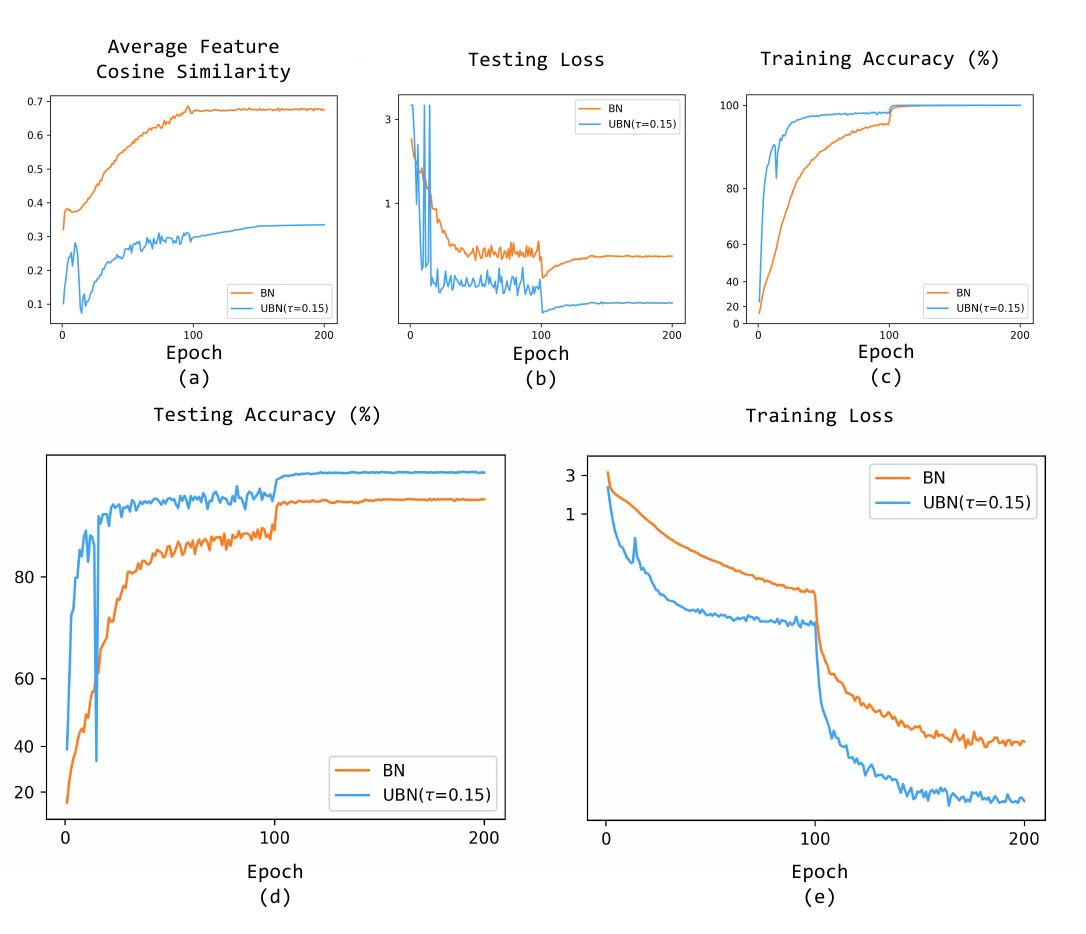}
    \captionsetup{skip=10pt}
    \caption{Feature condensation curves and learning curves of ResNet-50 on the CIFAR-10 dataset with batch size 256.  (a) Average feature cosine similarity of a fixed batch of samples during training. (b) Testing loss (c) Training accuracy (d) Testing accuracy (e) Training loss}
    \label{fig_supp:batch_size_cifar10_resnet50_bs=256}
\end{figure}

\begin{figure}[htbp]
    \centering
    \includegraphics[width=0.6\linewidth]{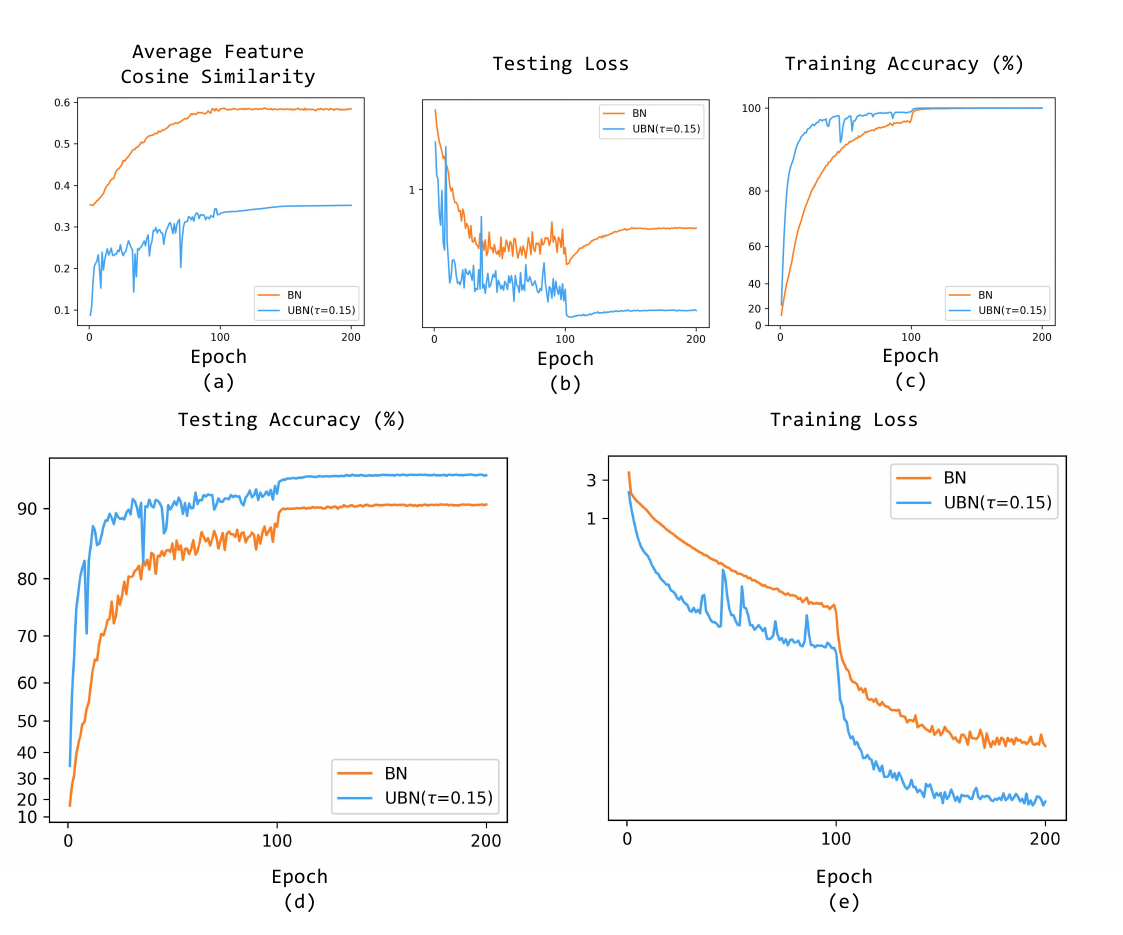}
    \captionsetup{skip=10pt}
    \caption{Feature condensation curves and learning curves of ResNet-50 on the CIFAR-10 dataset with batch size 512.  (a) Average feature cosine similarity of a fixed batch of samples during training. (b) Testing loss (c) Training accuracy (d) Testing accuracy (e) Training loss}
    \label{fig_supp:batch_size_cifar10_resnet50_bs=512}
\end{figure}

\begin{figure}[htbp]
    \centering
    \includegraphics[width=0.6\linewidth]{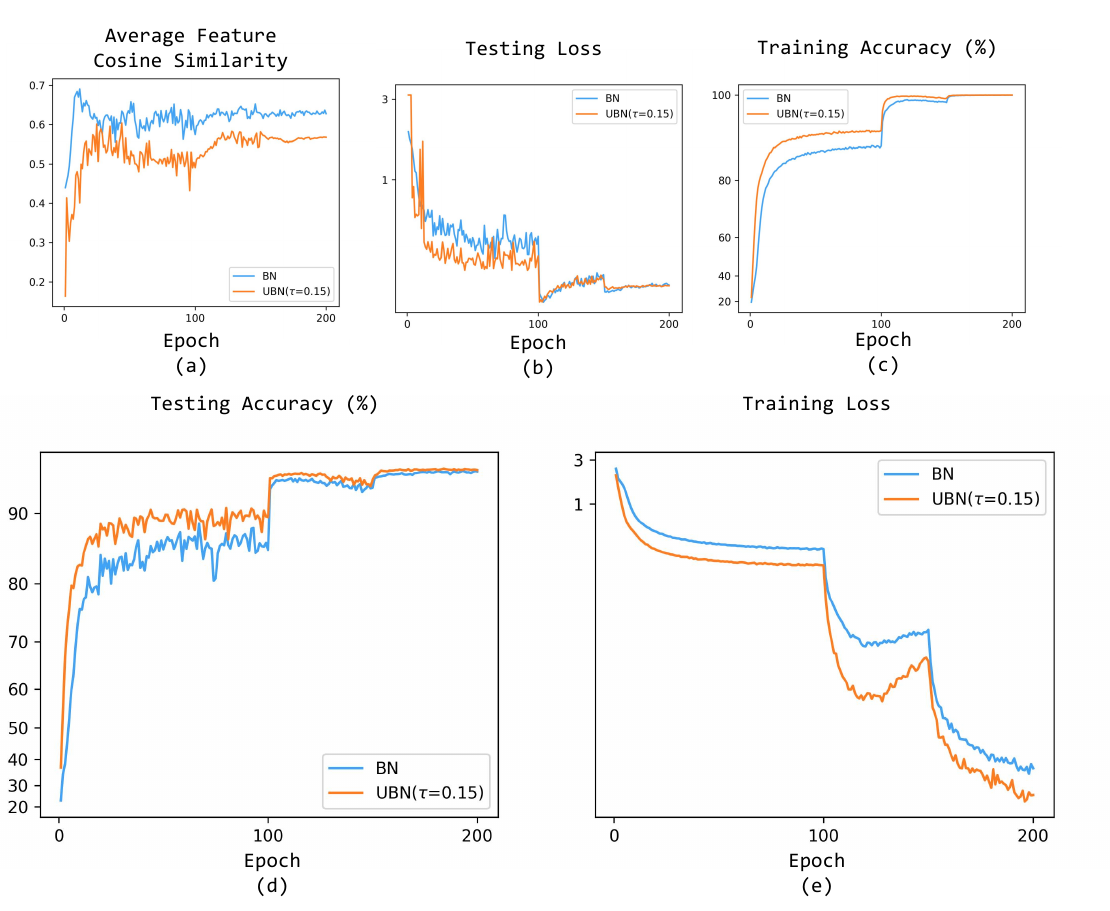}
    \captionsetup{skip=10pt}
    \caption{Feature condensation curves and learning curves of ResNet-101 on the CIFAR-10 dataset with batch size 32.  (a) Average feature cosine similarity of a fixed batch of samples during training. (b) Testing loss (c) Training accuracy (d) Testing accuracy (e) Training loss}
    \label{fig_supp:batch_size_cifar10_resnet101_bs=32}
\end{figure}

\begin{figure}[htbp]
    \centering
    \includegraphics[width=0.6\linewidth]{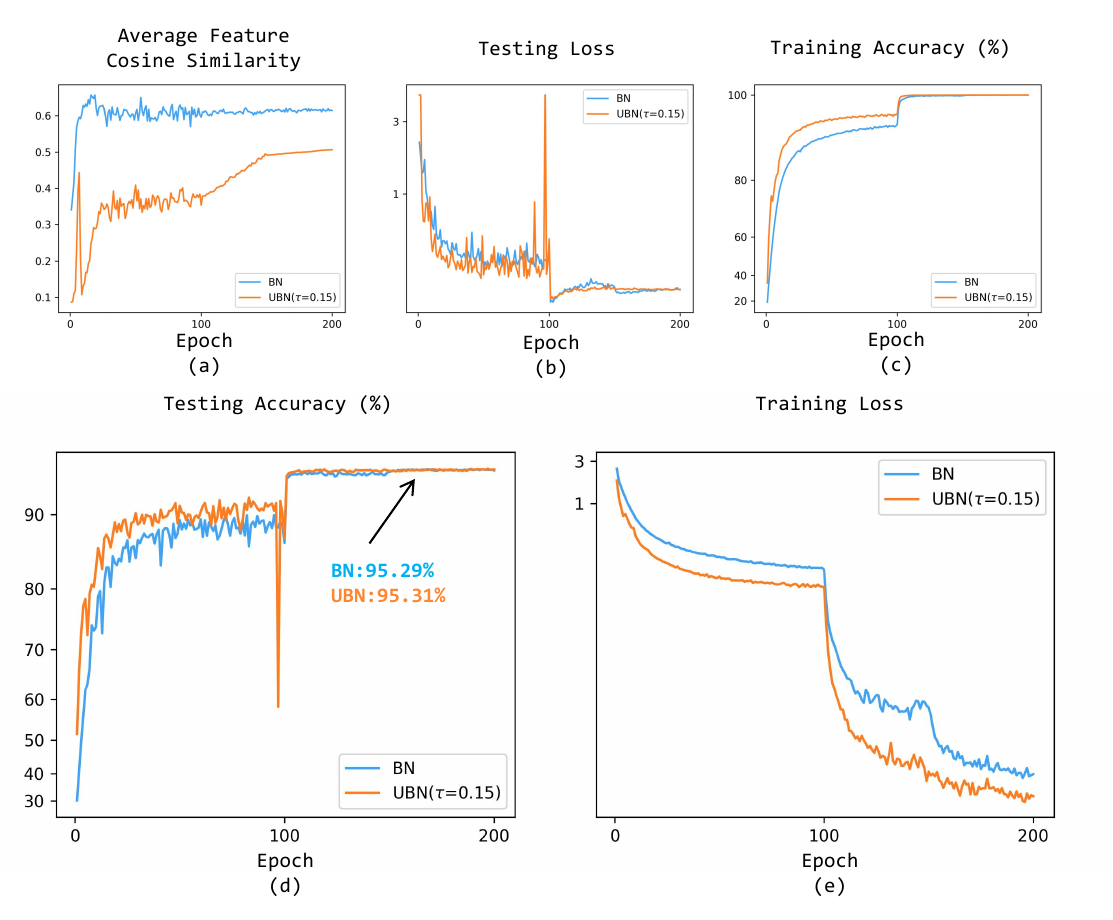}
    \captionsetup{skip=10pt}
    \caption{Feature condensation curves and learning curves of ResNet-101 on the CIFAR-10 dataset with batch size 64.  (a) Average feature cosine similarity of a fixed batch of samples during training. (b) Testing loss (c) Training accuracy (d) Testing accuracy (e) Training loss}
    \label{fig_supp:batch_size_cifar10_resnet101_bs=64}
\end{figure}

\begin{figure}[htbp]
    \centering
    \includegraphics[width=0.55\linewidth]{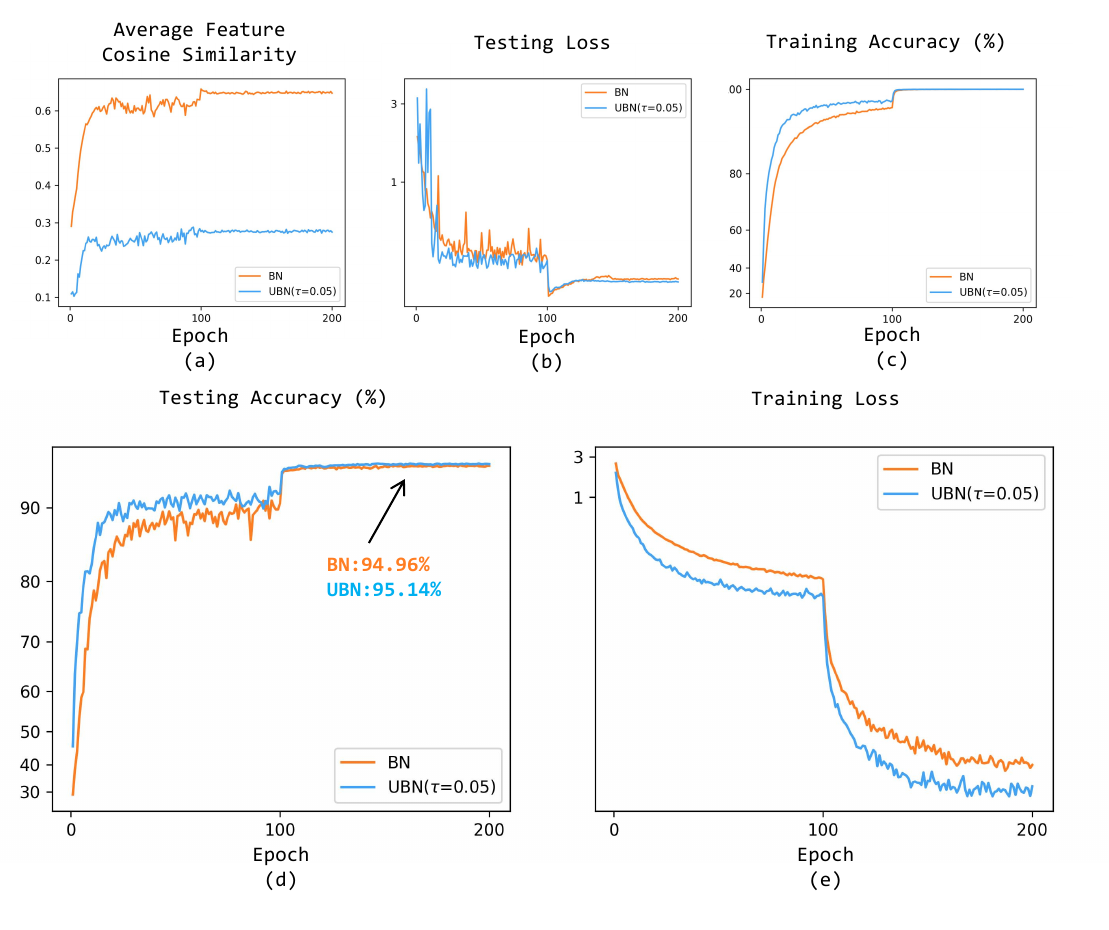}
    \captionsetup{skip=10pt}
    \caption{Feature condensation curves and learning curves of ResNet-101 on the CIFAR-10 dataset with batch size 128.  (a) Average feature cosine similarity of a fixed batch of samples during training. (b) Testing loss (c) Training accuracy (d) Testing accuracy (e) Training loss}
    \label{fig_supp:batch_size_cifar10_resnet101_bs=128}
\end{figure}

\begin{figure}[htbp]
    \centering
    \includegraphics[width=0.6\linewidth]{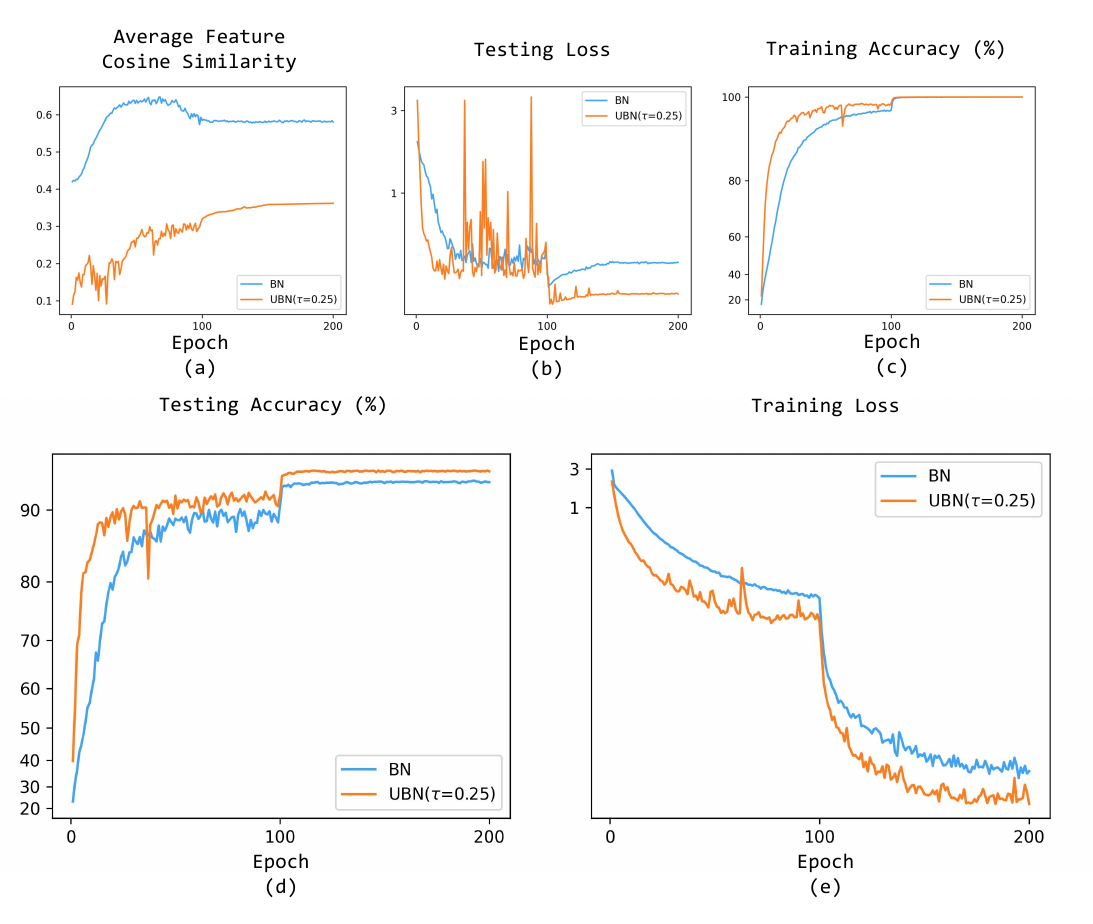}
    \captionsetup{skip=10pt}
    \caption{Feature condensation curves and learning curves of ResNet-101 on the CIFAR-10 dataset with batch size 256.  (a) Average feature cosine similarity of a fixed batch of samples during training. (b) Testing loss (c) Training accuracy (d) Testing accuracy (e) Training loss}
    \label{fig_supp:batch_size_cifar10_resnet101_bs=256}
\end{figure}

\begin{figure}[htbp]
    \centering
    \includegraphics[width=0.6\linewidth]{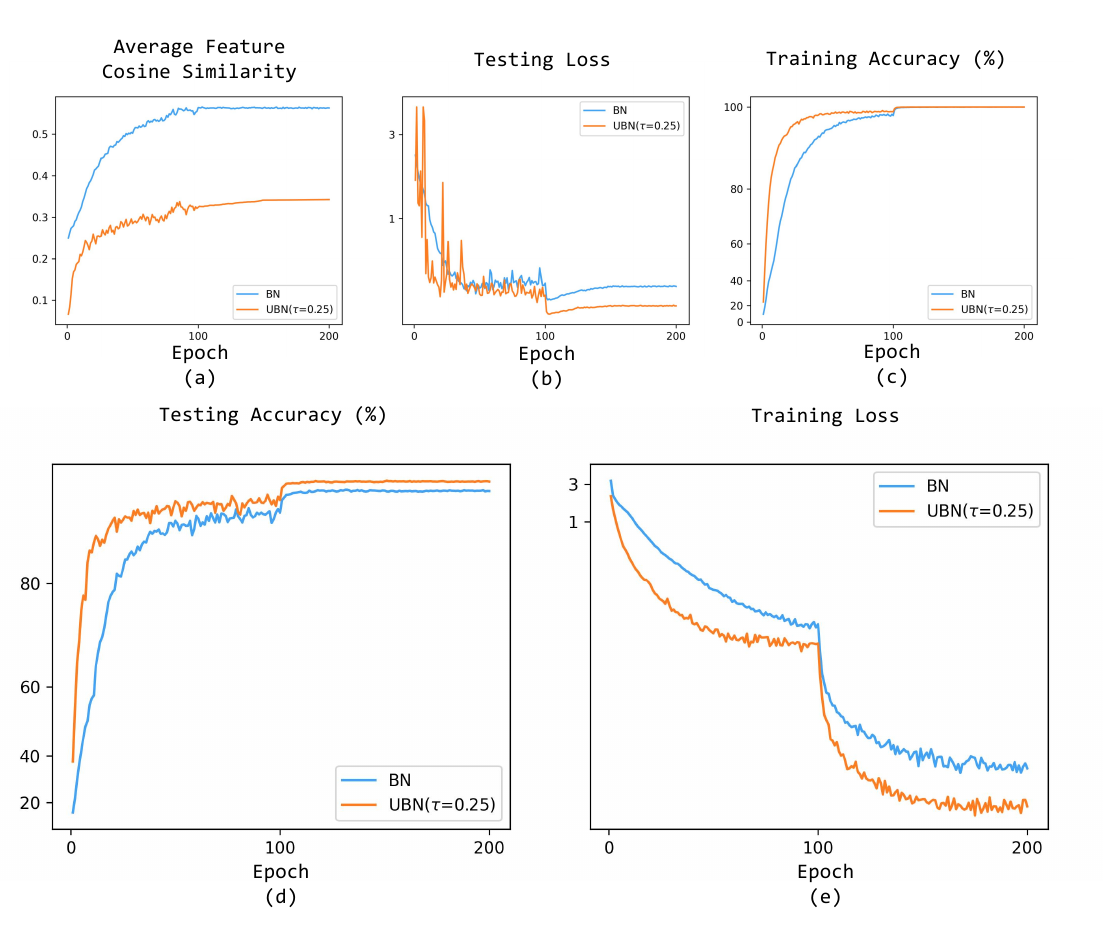}
    \captionsetup{skip=10pt}
    \caption{Feature condensation curves and learning curves of ResNet-101 on the CIFAR-10 dataset with batch size 512.  (a) Average feature cosine similarity of a fixed batch of samples during training. (b) Testing loss (c) Training accuracy (d) Testing accuracy (e) Training loss}
    \label{fig_supp:batch_size_cifar10_resnet101_bs=512}
\end{figure}

\begin{figure}[htbp]
    \centering
    \vspace{-20pt}
    \includegraphics[width=0.6\linewidth]{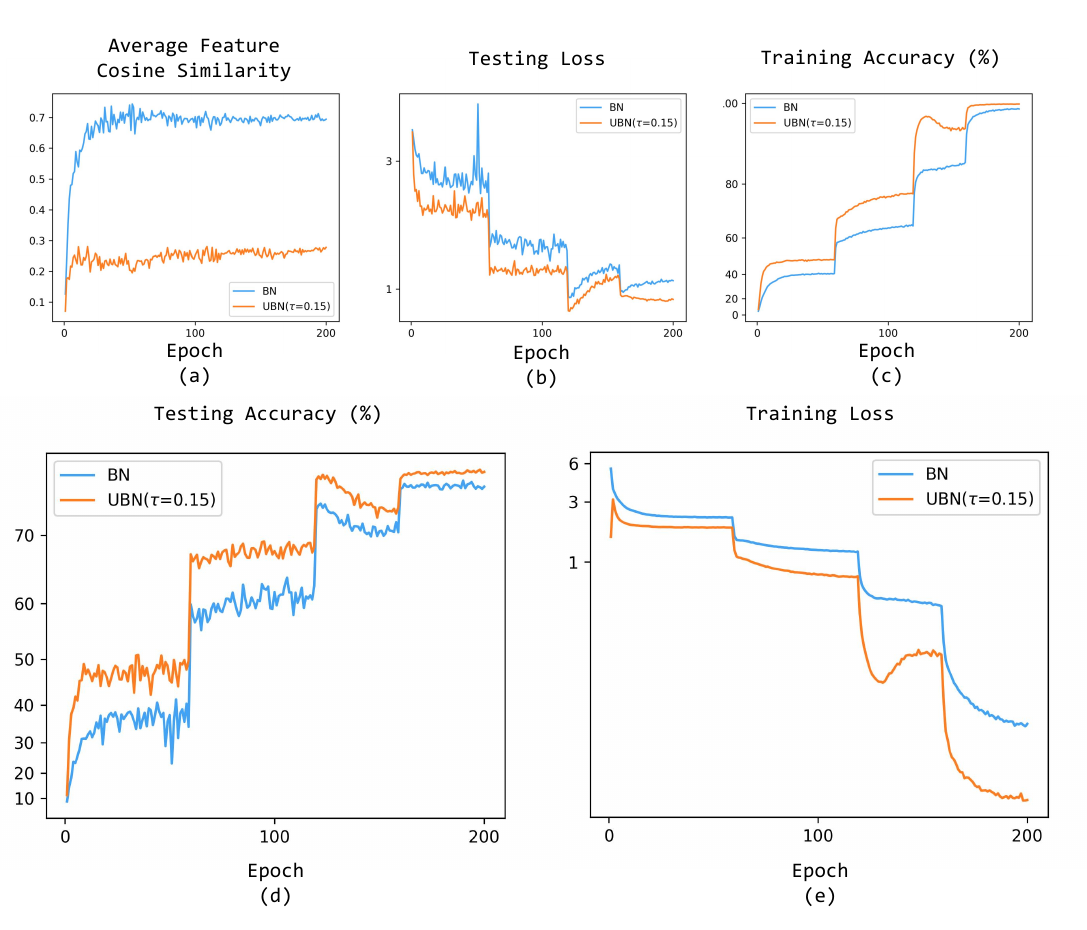}
    \captionsetup{skip=10pt}
    \caption{Feature condensation curves and learning curves of ResNet-50 on the CIFAR-100 dataset with batch size 32.  (a) Average feature cosine similarity of a fixed batch of samples during training. (b) Testing loss (c) Training accuracy (d) Testing accuracy (e) Training loss}
    \vspace{-20pt}
    \label{fig_supp:batch_size_cifar100_resnet50_bs=32}
\end{figure}

\begin{figure}[htbp]
    \centering
    \vspace{-20pt}
    \includegraphics[width=0.6\linewidth]{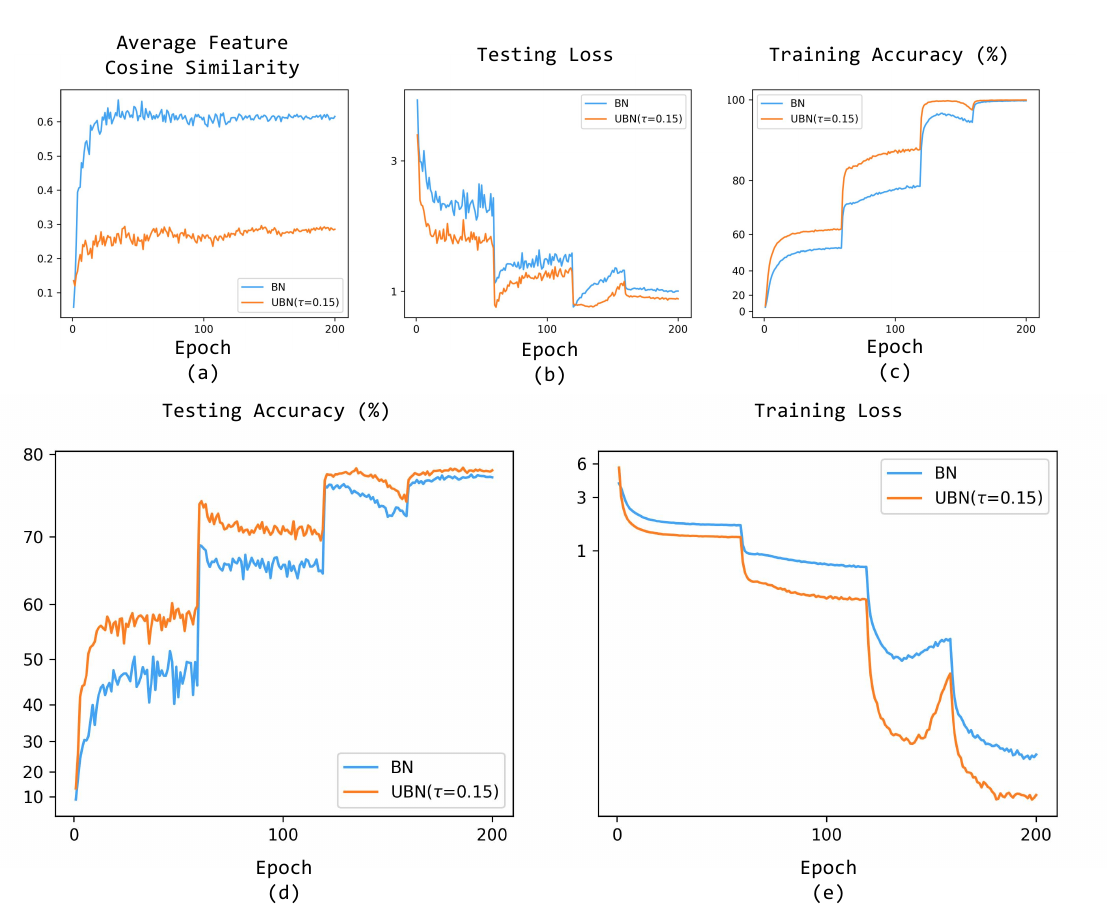}
    \captionsetup{skip=10pt}
    \caption{Feature condensation curves and learning curves of ResNet-50 on the CIFAR-100 dataset with batch size 64.  (a) Average feature cosine similarity of a fixed batch of samples during training. (b) Testing loss (c) Training accuracy (d) Testing accuracy (e) Training loss}
    \vspace{-20pt}
    \label{fig_supp:batch_size_cifar100_resnet50_bs=64}
\end{figure}

\begin{figure}[htbp]
    \centering
    \includegraphics[width=0.6\linewidth]{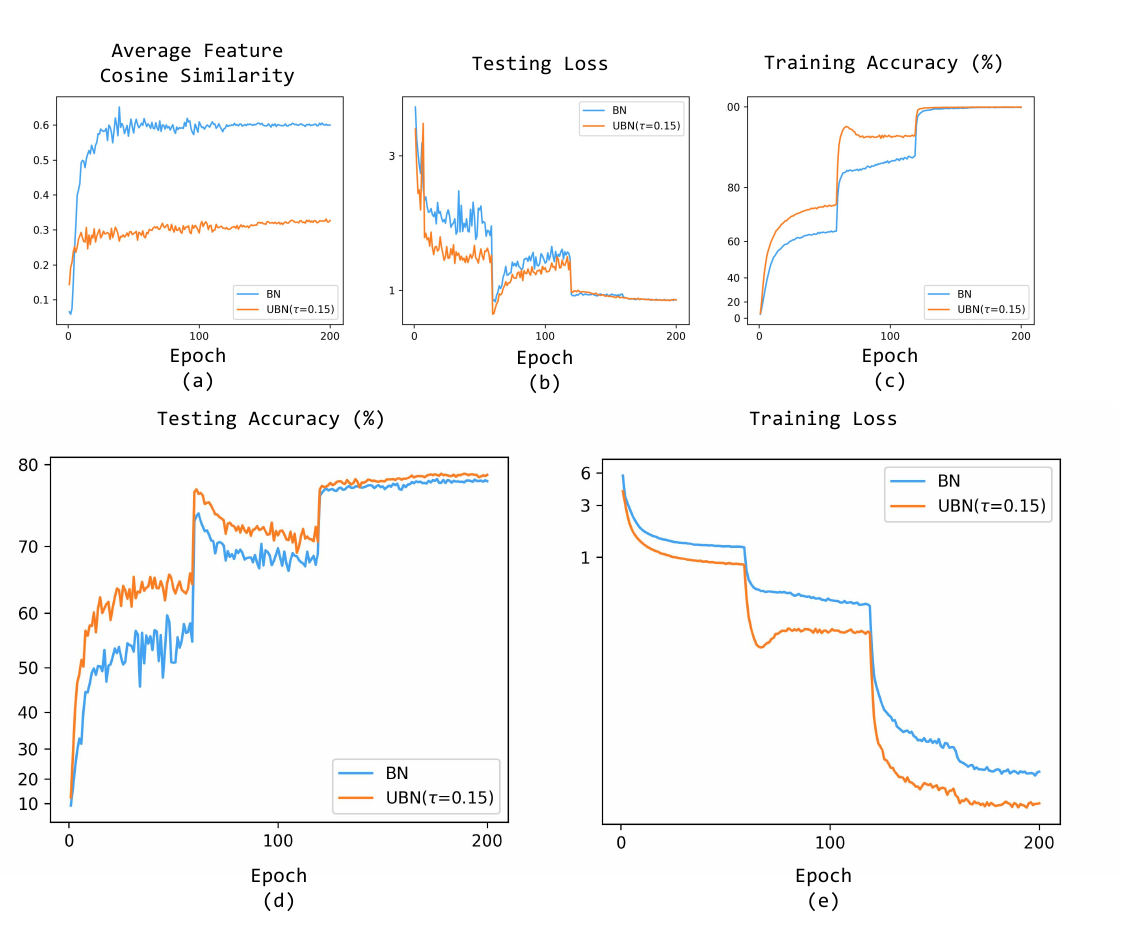}
    \captionsetup{skip=10pt}
    \caption{Feature condensation curves and learning curves of ResNet-50 on the CIFAR-100 dataset with batch size 128.  (a) Average feature cosine similarity of a fixed batch of samples during training. (b) Testing loss (c) Training accuracy (d) Testing accuracy (e) Training loss}
    \label{fig_supp:batch_size_cifar100_resnet50_bs=128}
\end{figure}

\begin{figure}[htbp]
    \centering
    \includegraphics[width=0.6\linewidth]{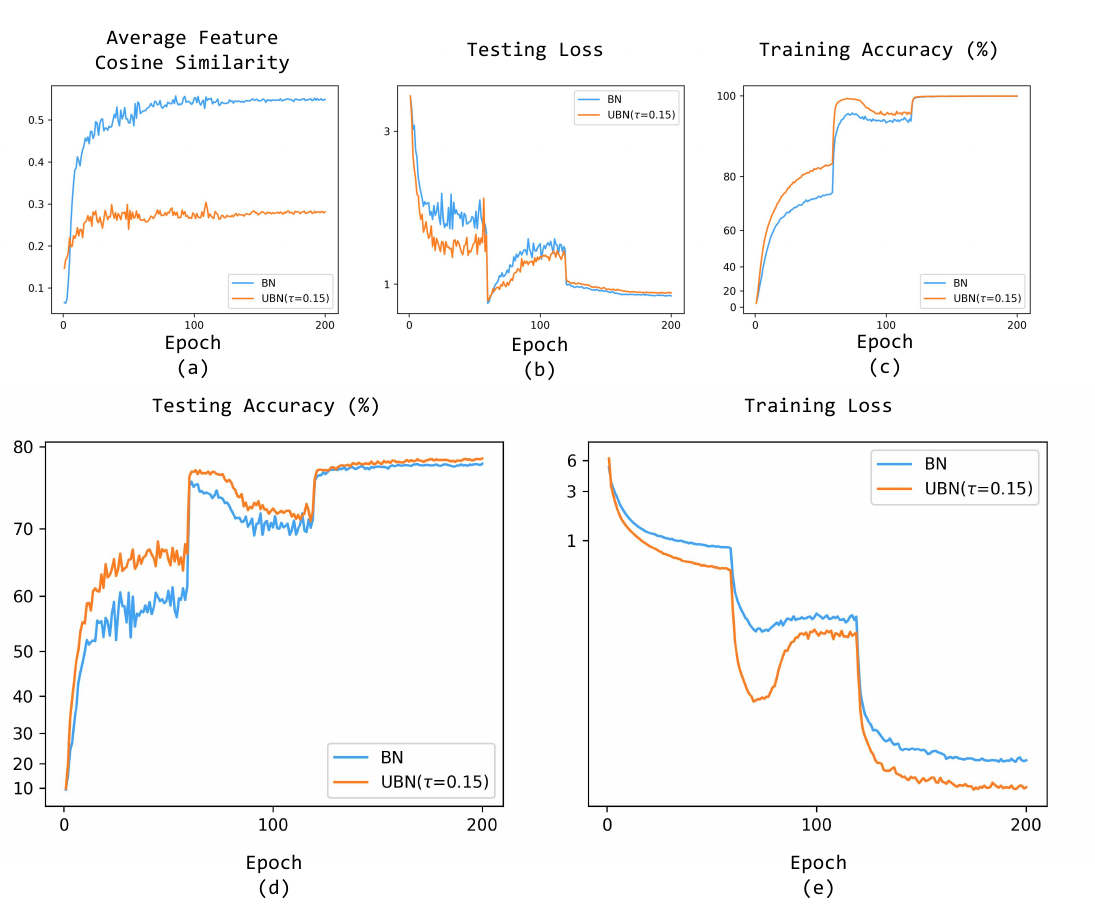}
    \captionsetup{skip=10pt}
    \caption{Feature condensation curves and learning curves of ResNet-50 on the CIFAR-100 dataset with batch size 256.  (a) Average feature cosine similarity of a fixed batch of samples during training. (b) Testing loss (c) Training accuracy (d) Testing accuracy (e) Training loss}
    \label{fig_supp:batch_size_cifar100_resnet50_bs=256}
\end{figure}

\begin{figure}[htbp]
    \centering
    \includegraphics[width=0.6\linewidth]{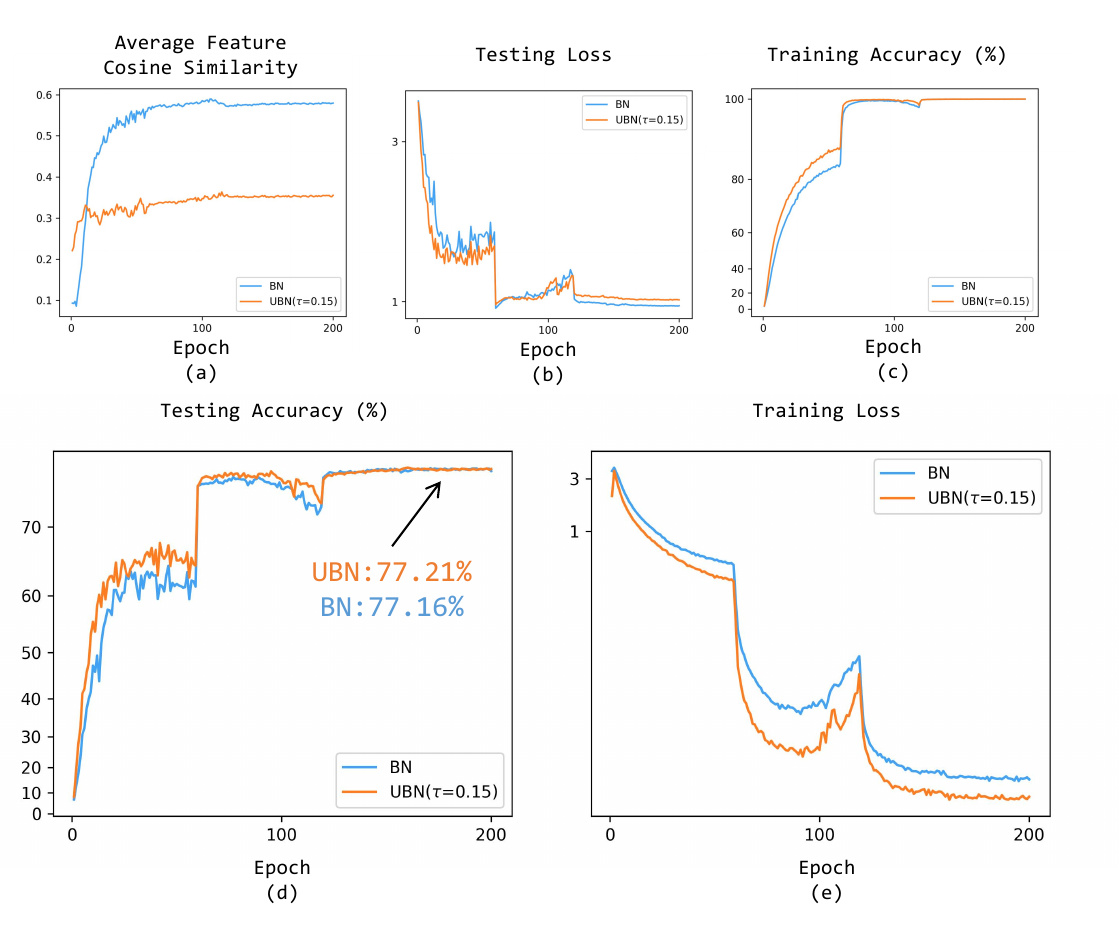}
    \captionsetup{skip=10pt}
    \caption{Feature condensation curves and learning curves of ResNet-50 on the CIFAR-100 dataset with batch size 512.  (a) Average feature cosine similarity of a fixed batch of samples during training. (b) Testing loss (c) Training accuracy (d) Testing accuracy (e) Training loss}
    \label{fig_supp:batch_size_cifar100_resnet50_bs=512}
\end{figure}

\begin{figure}[htbp]
    \centering
    \includegraphics[width=0.6\linewidth]{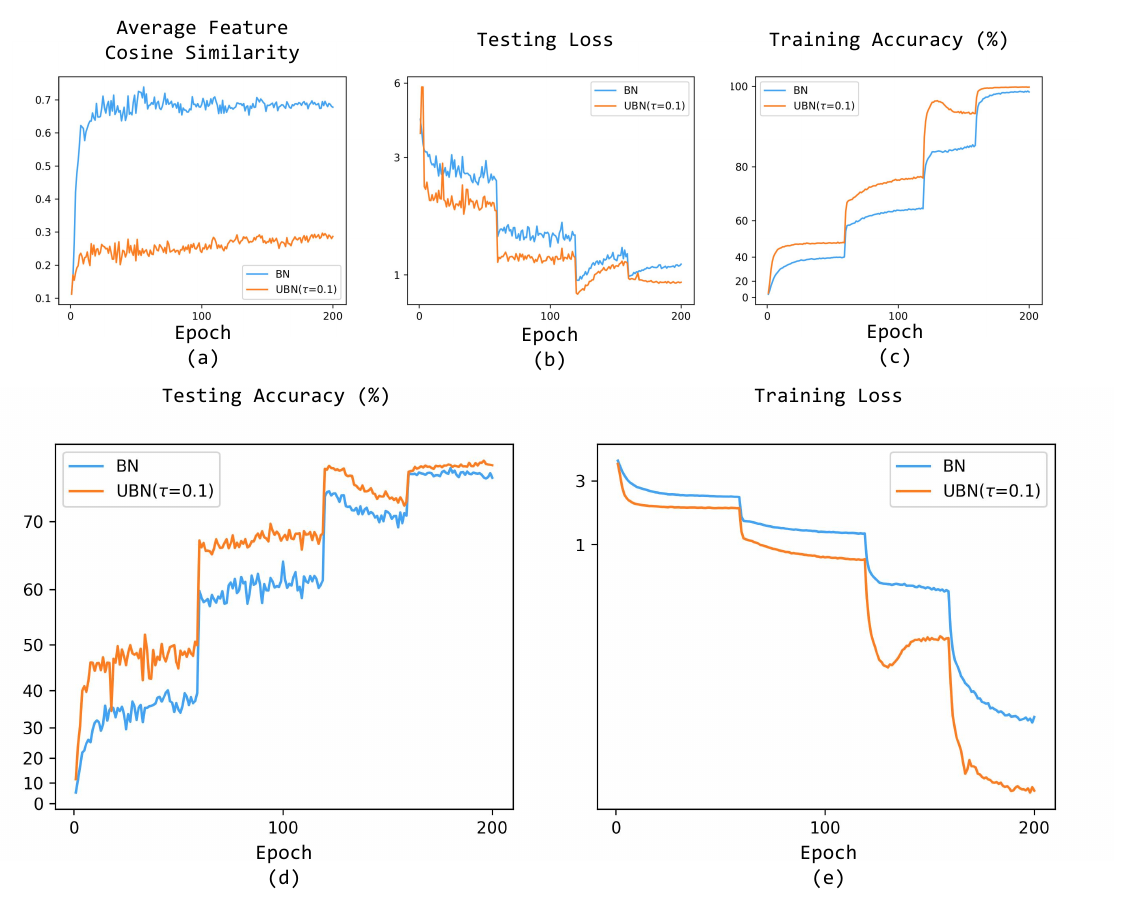}
    \captionsetup{skip=10pt}
    \caption{Feature condensation curves and learning curves of ResNet-101 on the CIFAR-100 dataset with batch size 32.  (a) Average feature cosine similarity of a fixed batch of samples during training. (b) Testing loss (c) Training accuracy (d) Testing accuracy (e) Training loss}
    \label{fig_supp:batch_size_cifar100_resnet101_bs=32}
\end{figure}

\begin{figure}[htbp]
    \centering
    \includegraphics[width=0.6\linewidth]{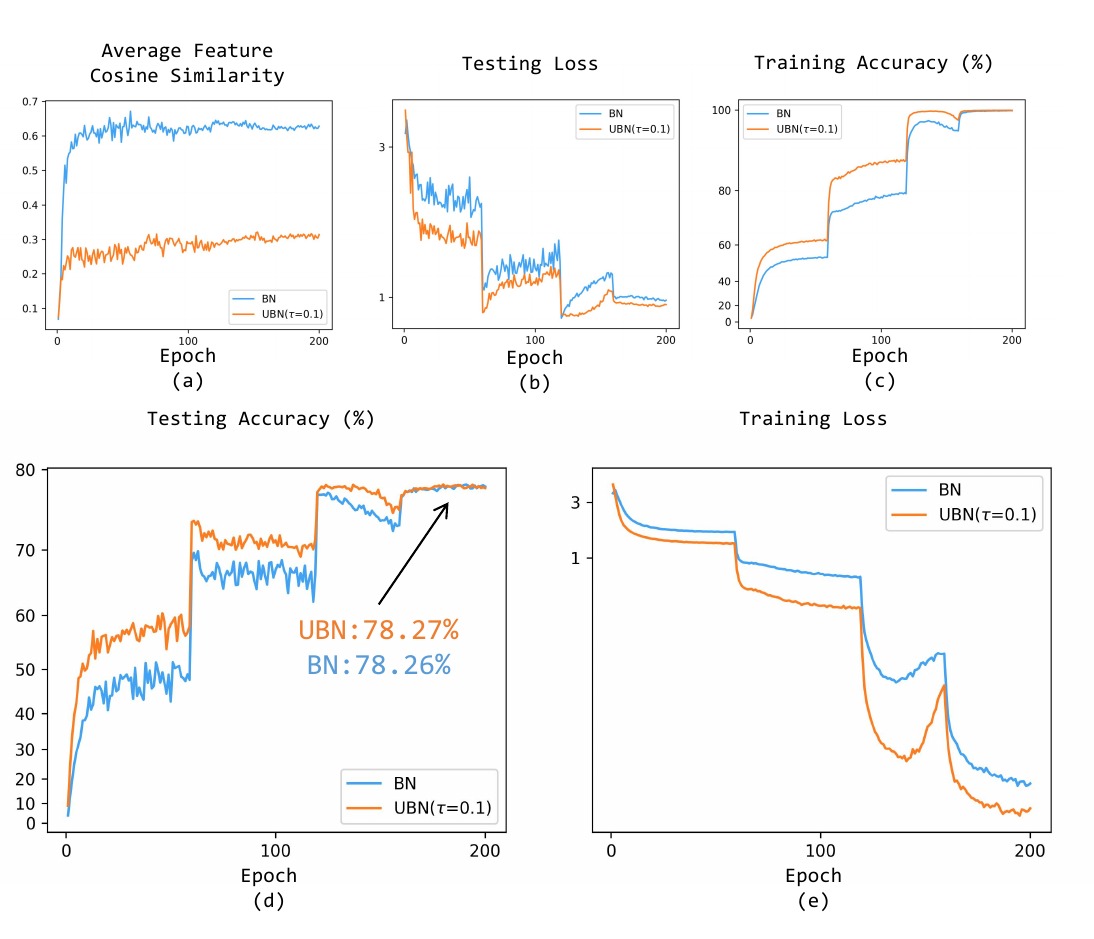}
    \captionsetup{skip=10pt}
    \caption{Feature condensation curves and learning curves of ResNet-101 on the CIFAR-100 dataset with batch size 64.  (a) Average feature cosine similarity of a fixed batch of samples during training. (b) Testing loss (c) Training accuracy (d) Testing accuracy (e) Training loss}
    \label{fig_supp:batch_size_cifar100_resnet101_bs=64}
\end{figure}

\begin{figure}[htbp]
    \centering
    \includegraphics[width=0.6\linewidth]{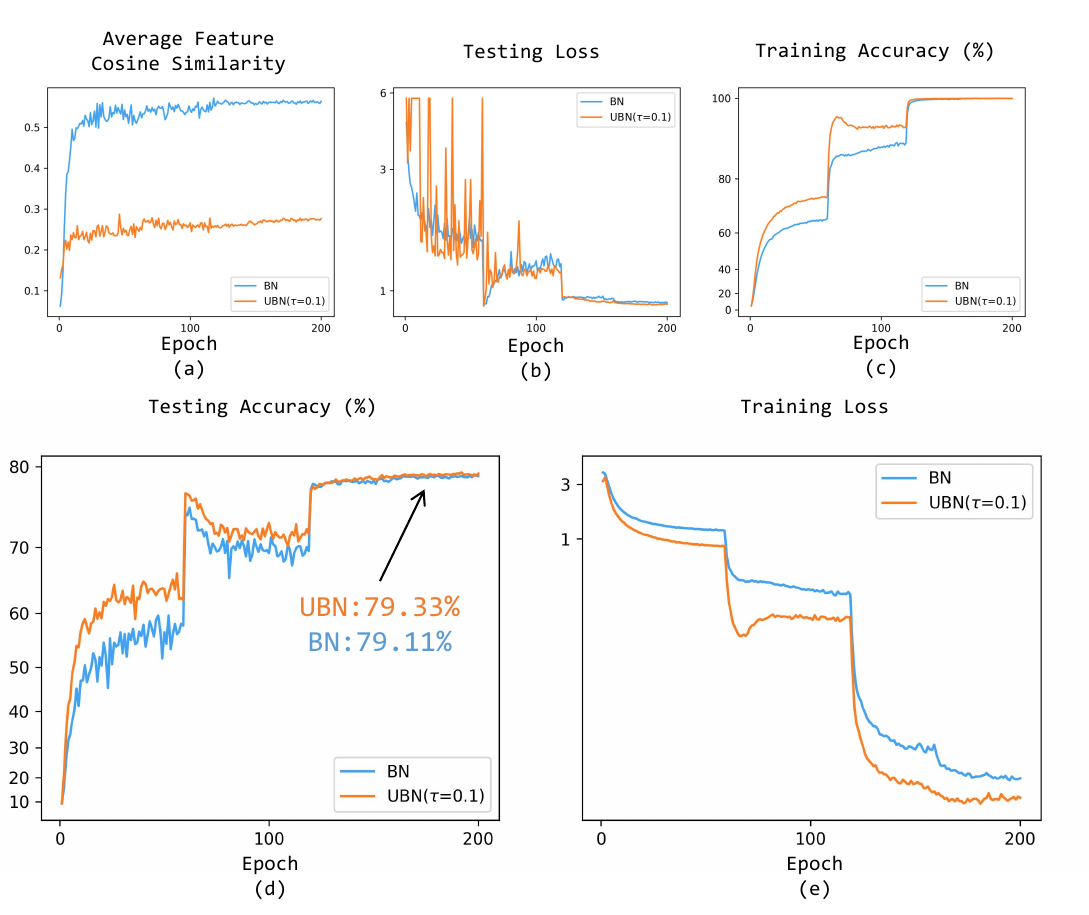}
    \captionsetup{skip=10pt}
    \caption{Feature condensation curves and learning curves of ResNet-101 on the CIFAR-100 dataset with batch size 128.  (a) Average feature cosine similarity of a fixed batch of samples during training. (b) Testing loss (c) Training accuracy (d) Testing accuracy (e) Training loss}
    \label{fig_supp:batch_size_cifar100_resnet101_bs=128}
\end{figure}

\begin{figure}[htbp]
    \centering
    \includegraphics[width=0.55\linewidth]{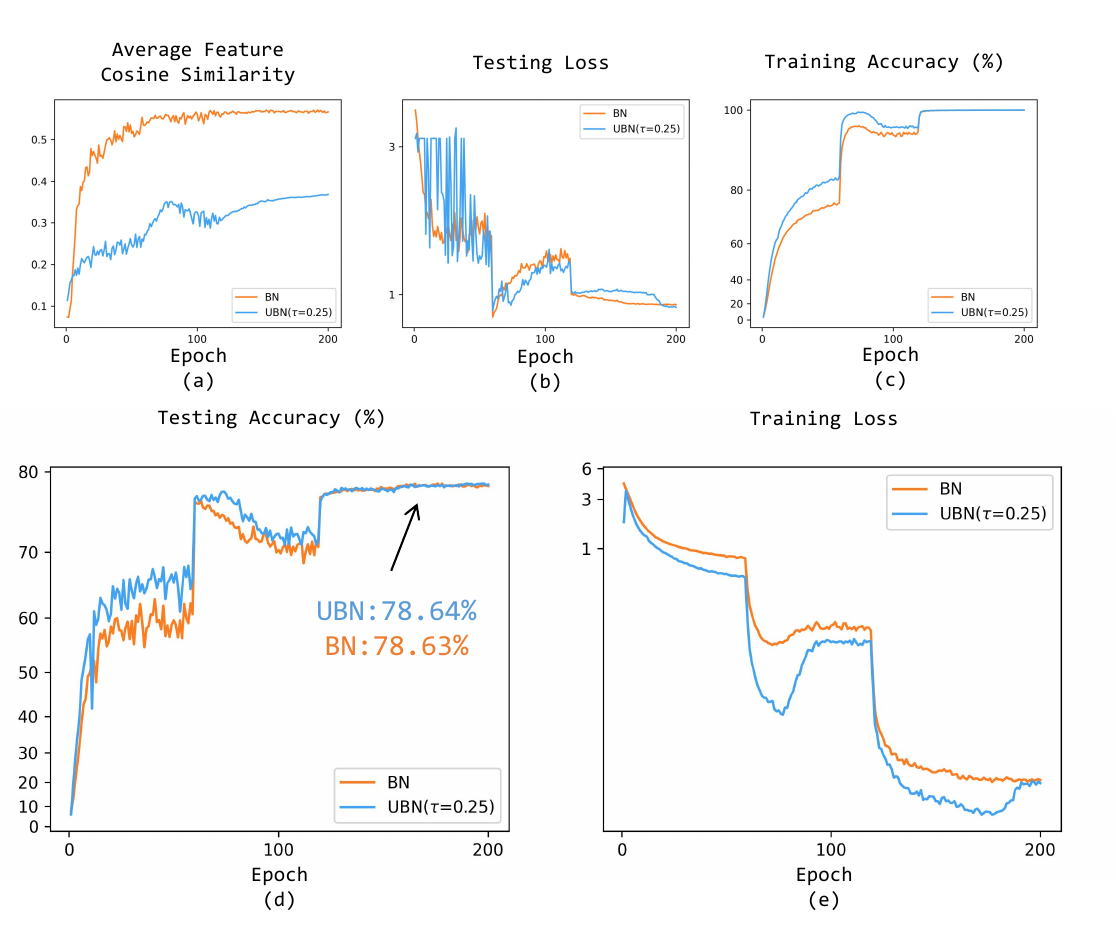}
    \captionsetup{skip=10pt}
    \caption{Feature condensation curves and learning curves of ResNet-101 on the CIFAR-100 dataset with batch size 256.  (a) Average feature cosine similarity of a fixed batch of samples during training. (b) Testing loss (c) Training accuracy (d) Testing accuracy (e) Training loss}
    \label{fig_supp:batch_size_cifar100_resnet101_bs=256}
\end{figure}

\begin{figure}[htbp]
    \centering
    \includegraphics[width=0.6\linewidth]{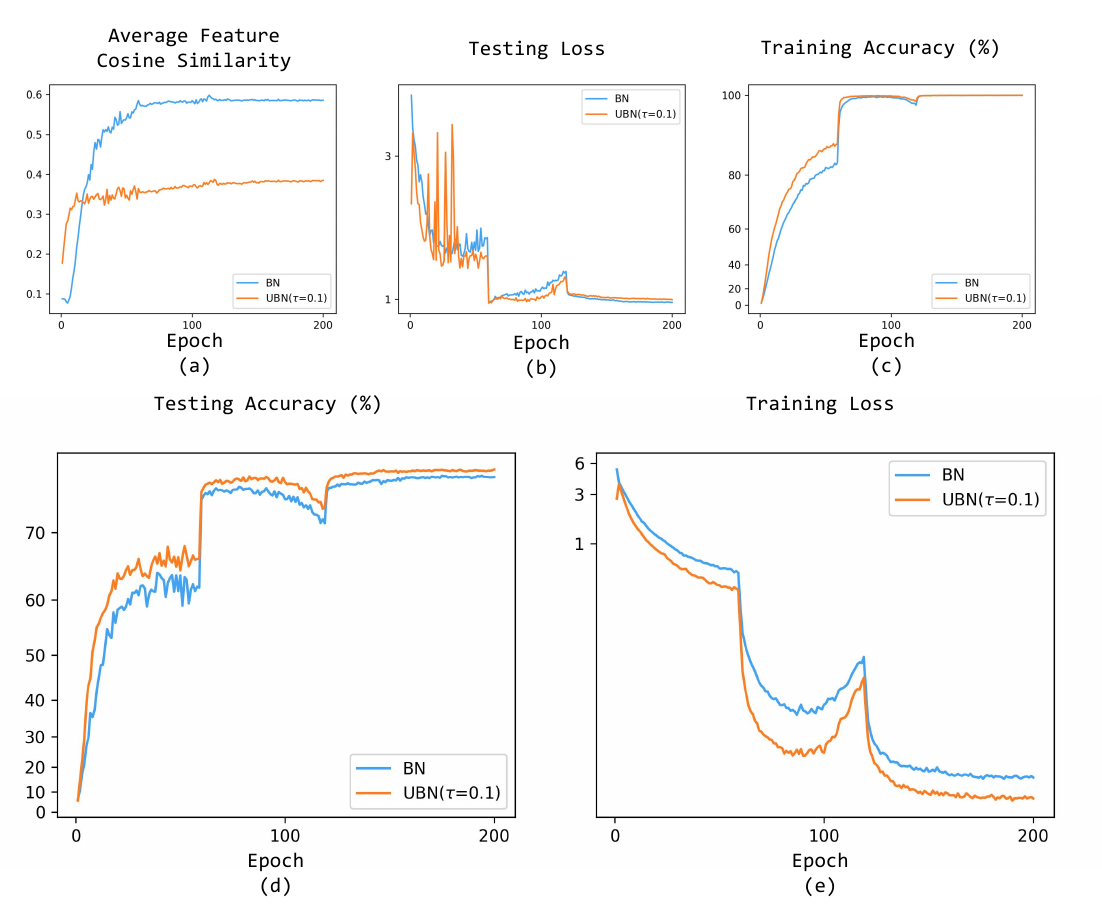}
    \captionsetup{skip=10pt}
    \caption{Feature condensation curves and learning curves of ResNet-101 on the CIFAR-100 dataset with batch size 512.  (a) Average feature cosine similarity of a fixed batch of samples during training. (b) Testing loss (c) Training accuracy (d) Testing accuracy (e) Training loss}
    \label{fig_supp:batch_size_cifar100_resnet101_bs=512}
\end{figure}

In this section, we conducted more experiments on different datasets and different models to compare the effect of different batch sizes in the training of UBN compared with traditional BN, which is shown in Figure~\ref{fig:ablation-bs} before. We conducted experiments on the ResNet-50 and ResNet-101, with batch size $32, 64, 128, 256, 512$, respectively. Extensive results of ResNet-50 on CIFAR-10 dataset are shown in Figure~\ref{fig_supp:batch_size_cifar10_resnet50_bs=32}, Figure~\ref{fig_supp:batch_size_cifar10_resnet50_bs=64}, Figure~\ref{fig_supp:batch_size_cifar10_resnet50_bs=128}, Figure~\ref{fig_supp:batch_size_cifar10_resnet50_bs=256}, and Figure~\ref{fig_supp:batch_size_cifar10_resnet50_bs=512}. Extensive results of ResNet-101 on CIFAR-10 dataset are shown in Figure~\ref{fig_supp:batch_size_cifar10_resnet101_bs=32}, Figure~\ref{fig_supp:batch_size_cifar10_resnet101_bs=64}, 
Figure~\ref{fig_supp:batch_size_cifar10_resnet101_bs=128}, 
Figure~\ref{fig_supp:batch_size_cifar10_resnet101_bs=256}, and Figure~\ref{fig_supp:batch_size_cifar10_resnet101_bs=512}. Extensive results of ResNet-50 on CIFAR-100 dataset are shown in Figure~\ref{fig_supp:batch_size_cifar100_resnet50_bs=32}, Figure~\ref{fig_supp:batch_size_cifar100_resnet50_bs=64}, 
Figure~\ref{fig_supp:batch_size_cifar100_resnet50_bs=128}, 
Figure~\ref{fig_supp:batch_size_cifar100_resnet50_bs=256}, and Figure~\ref{fig_supp:batch_size_cifar100_resnet50_bs=512}. Extensive results of ResNet-101 on CIFAR-100 dataset are shown in Figure~\ref{fig_supp:batch_size_cifar100_resnet101_bs=32}, Figure~\ref{fig_supp:batch_size_cifar100_resnet101_bs=64}, Figure~\ref{fig_supp:batch_size_cifar100_resnet101_bs=128}, 
Figure~\ref{fig_supp:batch_size_cifar100_resnet101_bs=256}, 
and Figure~\ref{fig_supp:batch_size_cifar100_resnet101_bs=512}.

Our experiments were conducted on the CIFAR-10 and CIFAR-100 datasets, utilizing single Nividia A100 GPU. We followed the identical training settings outlined in Section ~\ref{sec:exp}.
For CIFAR-10 dataset, each model was trained for a total of 200 epochs. For learning rate management, we applied a multi-step decay strategy. Our optimization choice is SGD, starting with a learning rate of 0.1, a momentum of 0.9, and a weight decay rate of 1e-4. We programmed the learning rate to reduce by 0.1 at the 100th and 150th epochs.
For CIFAR-100 dataset, each model underwent 200 training epochs. We introduced a warm-up phase in the first epoch, where the learning rate increased linearly with each batch to stabilize early training. After this, we employed a multi-step decay for the learning rate. Using SGD as the optimizer, we started with an initial learning rate of 0.1, a momentum of 0.9, and a weight decay of 5e-4. The learning rate was set to decrease by 0.2 at the 60th, 120th, and 160th epochs.

\subsection{On the CIFAR-10 dataset}

\clearpage
\subsection{On the CIFAR-100 dataset}

\clearpage

\section{Exploring the Functionality of Rectifications for UBN in Additional Experiments}
\label{sec:supp_ablation}
\begin{figure}[htbp]
    \centering
    \includegraphics[width=0.55\linewidth]{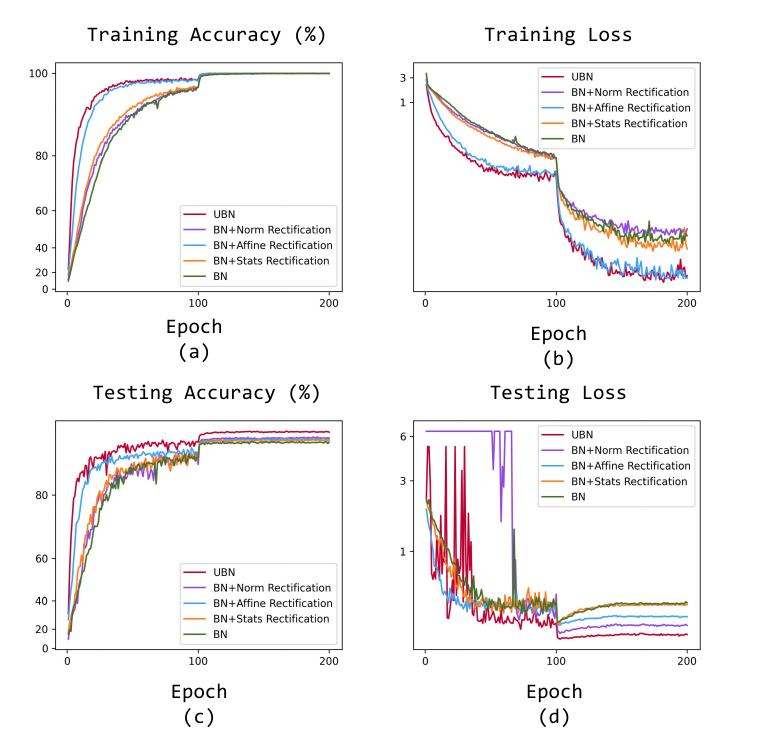}
    \captionsetup{skip=10pt}
    \caption{Learning curves of ResNet-50 on the CIFAR-10. (a) Training accuracy (b) Training loss (c) Testing accuracy (d) Testing loss   }
    \label{fig_supp:bntype_cifar10_resnet50}
\end{figure}

\begin{figure}[htbp]
    \centering
    \includegraphics[width=0.55\linewidth]{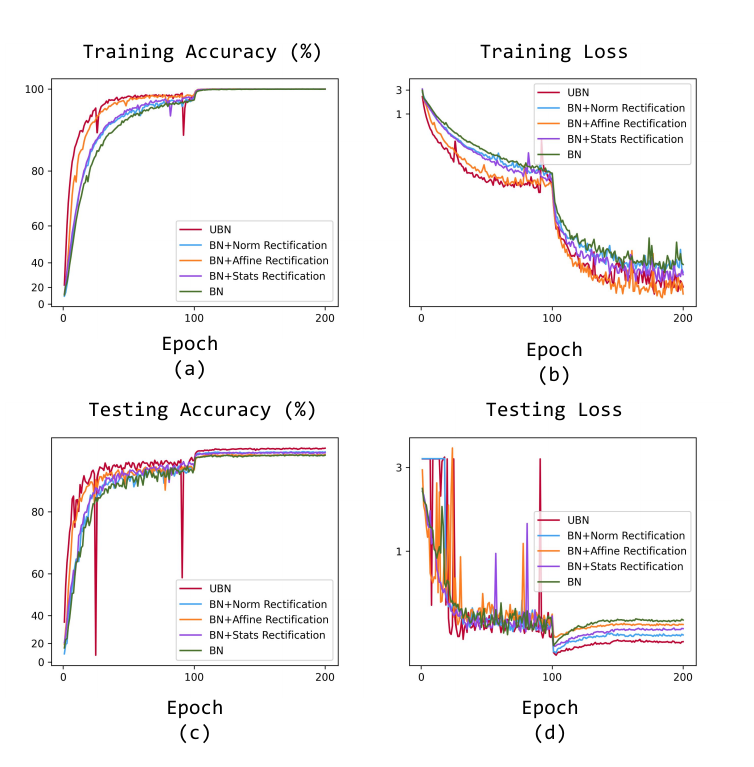}
    \captionsetup{skip=10pt}
    \caption{Learning curves of ResNet-101 on the CIFAR-10. (a) Training accuracy (b) Training loss (c) Testing accuracy (d) Testing loss   }
    \vspace{-20pt}
    \label{fig_supp:bntype_cifar10_resnet101}
\end{figure}

\begin{figure}[htbp]
    \centering
    \includegraphics[width=0.55\linewidth]{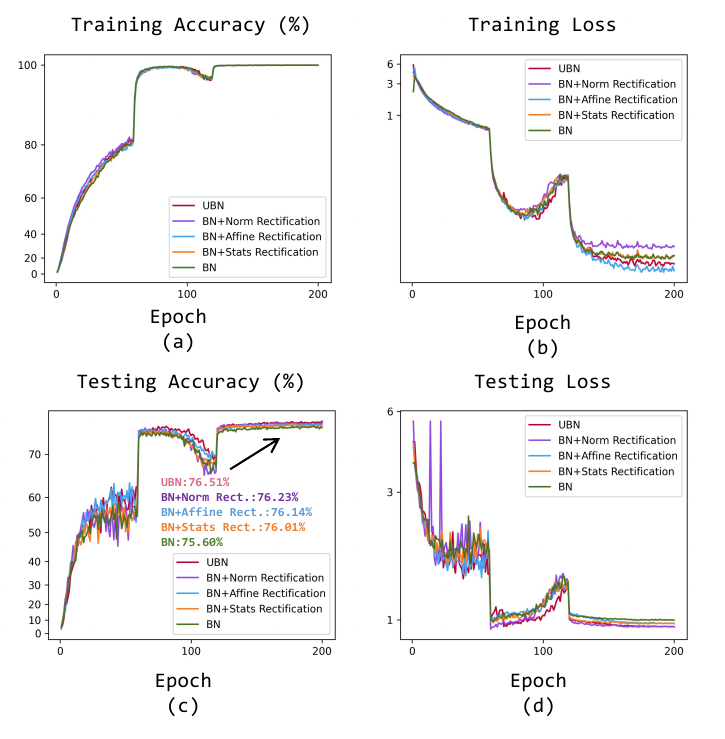}
    \captionsetup{skip=3pt}
    \caption{Learning curves of Inceptionv4 on the CIFAR-100. (a) Training accuracy (b) Training loss (c) Testing accuracy (d) Testing loss   }
    \label{fig_supp:bntype_cifar100_inceptionv4}
\end{figure}

\begin{figure}[htbp]
    \centering
    \includegraphics[width=0.55\linewidth]{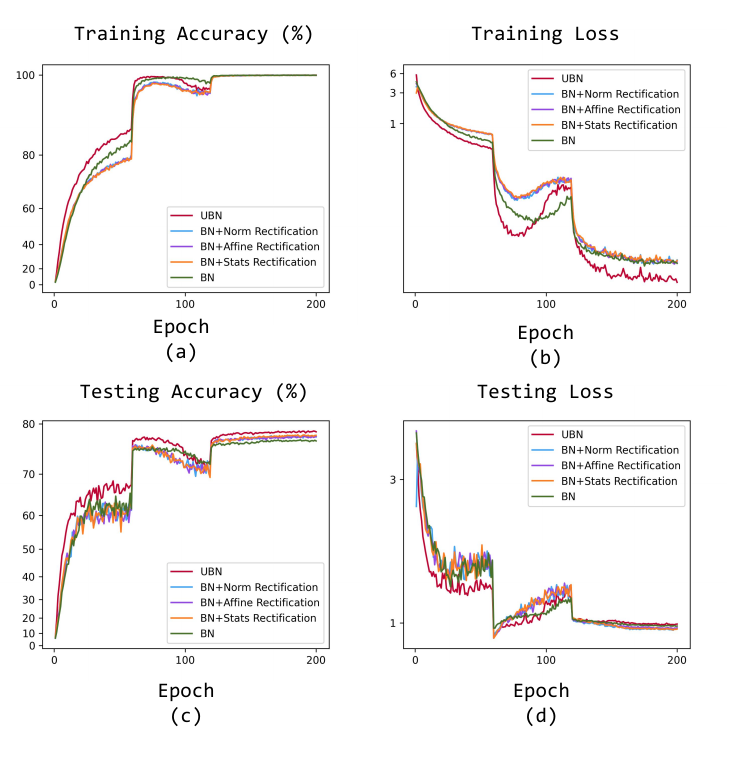}
    \captionsetup{skip=3pt}
    \caption{Learning curves of ResNet-50 on the CIFAR-100. (a) Training accuracy (b) Training loss (c) Testing accuracy (d) Testing loss   }
    \label{fig_supp:bntype_cifar100_resnet50}
\end{figure}

In this section, we conducted more experiments on different datasets and different models to perform ablation study on different rectifications in the training of UBN, which is shown in Figure~\ref{fig:ablation-rec} before. Extensive results on CIFAR-10 dataset are shown in Figure~\ref{fig_supp:bntype_cifar10_resnet50} and Figure~\ref{fig_supp:bntype_cifar10_resnet101}. Extensive results on CIFAR-100 dataset are shown in Figure~\ref{fig_supp:bntype_cifar100_inceptionv4} and Figure~\ref{fig_supp:bntype_cifar100_resnet50}. 

Our experiments were conducted on the CIFAR-10 and CIFAR-100 datasets, utilizing three Nvidia 3090 GPUs. We followed the identical training settings outlined in Section ~\ref{sec:exp}.
For CIFAR-10 dataset, each model was trained for a total of 200 epochs. For learning rate management, we applied a multi-step decay strategy. Our optimization choice is SGD, starting with a learning rate of 0.1, a momentum of 0.9, and a weight decay rate of 1e-4. We programmed the learning rate to reduce by 0.1 at the 100th and 150th epochs. Each GPU was assigned a batch size of 128.
For CIFAR-100 dataset, each model underwent 200 training epochs. We introduced a warm-up phase in the first epoch, where the learning rate increased linearly with each batch to stabilize early training. After this, we employed a multi-step decay for the learning rate. Using SGD as the optimizer, we started with an initial learning rate of 0.1, a momentum of 0.9, and a weight decay of 5e-4. The learning rate was set to decrease by 0.2 at the 60th, 120th, and 160th epochs. We maintained a batch size of 128 per GPU.

\vspace{-20pt}

\vspace{-10pt}

\clearpage
\section{Analysis of training complexity} 
We've conducted new experiments to evaluate UBN's training efficiency by calculating the comparative training time when acheving the comparative training performance for each model. The reported UBN based model results achieve similar performance levels to BN based model. Although determination of a feature condensation threshold may cause additional computational cost and time, UBN reaches comparable performance in less time, as shown in Table \ref{table:time}.

\begin{table}[h!]
\caption{Training time of BN and UBN on different models on CIFAR-10 and ImageNet datasets.}
\label{table:time}
\centering
    \begin{tabular}{l|lcc}
    \hline
    Dataset & Model &  Training Time (s)&  \#Epoch\\
    \hline
    \multirow{6}{*}{CIFAR-10} & ResNet-34-BN & 3760 & 200 \\
    &  ResNet-34-UBN ({\small $\tau=0.15$})&  3605 & 101 \\ \cline{2-4}
     & ResNet-50-BN & 4302 & 200 \\
    &  ResNet-50-UBN ({\small $\tau=0.15$})&  3606 & 73 \\ \cline{2-4}
         & VGG-11-BN & 1475 & 200 \\ 
    &  VGG-11-UBN ({\small $\tau=0.25$})&  1044 & 114 \\
    \hline
    \multirow{6}{*}{ImageNet} & Res2Net-50-BN & 33048 & 90 \\
    &  Res2Net-50-UBN ({\small $\tau=0.15$})& 23734 & 31 \\ \cline{2-4} 
    & ResNeXt-50-BN & 29367 & 90 \\
    &  ResNeXt-50-UBN ({\small $\tau=0.15$})& 22840 & 31 \\\cline{2-4}
    & ResNet-101-BN & 35594 & 90 \\
    &  ResNet-101-UBN ({\small $\tau=0.15$})& 26557 & 31 \\
    \hline
    \end{tabular}
\end{table}
\end{document}